\documentclass[10pt,journal,compsoc]{IEEEtran}

\usepackage{setspace}

\ifCLASSOPTIONcompsoc
  \usepackage[nocompress]{cite}
\else
  \usepackage{cite}
\fi
\usepackage{url}
\usepackage{times}
\usepackage{graphicx}

\usepackage{xcolor}
\usepackage{gensymb}
\usepackage{booktabs}
\usepackage{multirow}
\usepackage{ragged2e}
\usepackage{caption}
\usepackage{tabularx}
\usepackage[export]{adjustbox}
\usepackage{array}

\usepackage{amsfonts}
\usepackage{amsmath}
\usepackage{amssymb}

\usepackage{gensymb}
\usepackage{adjustbox}
\usepackage{pgfplots}
\pgfplotsset{compat=1.16}
\usepgfplotslibrary{statistics}
\usepackage{dirtytalk}
\usepackage{svg}
\usepackage{subcaption}

\usepackage{hyperref}
\hypersetup{
    colorlinks=true,
    linkcolor=blue,
    filecolor=magenta,      
    urlcolor=cyan,
}

\newcommand{\new}[1]{\textcolor{black}{{#1}}}

\begin{document}
\title{TRACK: A New Method from a Re-examination of Deep Architectures for Head Motion Prediction in 360\degree\ Videos}

\author{Miguel~Fabi\'an~Romero Rond\'on,
        Lucile~Sassatelli,
        Ram\'on~Aparicio Pardo,
        and~Fr\'ed\'eric~Precioso
\thanks{\new{A preliminary version of this work has been published in \cite{us_ICIP}}. The authors are with Universit\'e C\^ote d'Azur, CNRS, I3S, 06900 Sophia Antipolis, France. Lucile Sassatelli is also with Institut Universitaire de France. Fr\'ed\'eric Precioso is also with Inria. E-mail: \{first.last\}@univ-cotedazur.fr} }

\markboth{}
{Romero \MakeLowercase{\textit{et al.}}: TRACK: A New method from a Re-examination of Deep Architectures for Head Motion Prediction in 360\degree\ Videos}

\IEEEtitleabstractindextext{
\begin{abstract}
\justify
We consider predicting the user's head motion in 360\degree\ videos, with 2 modalities only: the past user's positions and the video content (not knowing other users' traces).
We make two main contributions. First, we re-examine existing deep-learning approaches for this problem and identify hidden flaws from a thorough root-cause analysis. Second, from the results of this analysis, we design a new proposal establishing state-of-the-art performance.
First, re-assessing the existing methods that use both modalities, we obtain the surprising result that they all perform worse than baselines using the user’s trajectory only. A root-cause analysis of the metrics, datasets and neural architectures shows in particular that (i) the content can inform the prediction for horizons longer than 2 to 3 sec. (existing methods consider shorter horizons), and that (ii) to compete with the baselines, it is necessary to have a recurrent unit dedicated to process the positions, but this is not sufficient. 
Second, from a re-examination of the problem supported with the concept of Structural-RNN, we design a new deep neural architecture, named TRACK. TRACK achieves state-of-the-art performance on all considered datasets and prediction horizons, outperforming competitors by up to 20\% on focus-type videos and horizons 2-5 seconds.
The entire framework (codes and datasets) is online and received an ACM reproducibility badge \url{https://gitlab.com/miguelfromeror/head-motion-prediction}.
\end{abstract}

}

\maketitle

\IEEEdisplaynontitleabstractindextext
\IEEEpeerreviewmaketitle

\section{Introduction}\label{sec:introduction}

\begin{figure}[h]
  \centering
  \includegraphics[width=1.0\columnwidth]{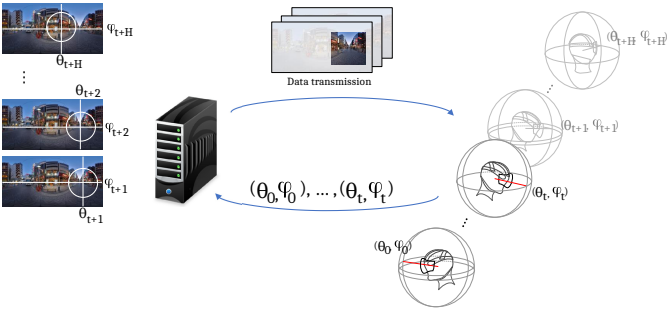}
  \caption{\textbf{360\degree\ video streaming principle}. The user requests the next video segment at time $t$, if the future orientations of the user $(\theta_{t+1}, \varphi_{t+1}), ..., (\theta_{t+H}, \varphi_{t+H})$ were known, the bandwidth consumption could be reduced by sending in higher quality only the areas corresponding to the future FoV.}
  \label{fig:headMotionPrediction}
\end{figure}

\IEEEPARstart{I}{mmersive} media are on the rise: 
the global market for Virtual Reality (VR) is projected to grow from US\$9.2 Billion in 2020 to US\$89.1 Billion by 2027 \cite{GIA2020}.
360\degree\ videos are an important modality of VR, with applications in story-telling, journalism or remote education.
Despite these exciting prospects, the development is persistently hindered by the difficulty to access immersive content through Internet streaming. Indeed, owing to the closer proximity of the screen to the eye in VR and to the width of the content ($2\pi$ steradians in azimuth and $\pi$ in elevation angles), the data rate is two orders of magnitude that of a regular video \cite{park2018}.
To decrease the amount of data to stream, a solution is to send in high resolution only the portion of the sphere the user has access to at each point in time, named the Field of View (FoV). To do so, recent works have proposed to either segment the video spatially into tiles and set the quality of the tiles according to their proximity to the FoV \cite{qian2016optimizing,xiao2018bas,gaddam2016tiling}, or use projections enabling high resolutions of regions close to the FoV \cite{hristovaheterogeneous,fu2009rhombic}.
These approaches however require to know the user's head position in advance, that is at the time of sending the content from the server (see Fig.~\ref{fig:headMotionPrediction}). Failing to predict correctly the future user's positions can lead to a lower quality displayed in the FoV, which can impair the user's experience.
It is therefore crucial for an efficient 360\degree\ video streaming system to embed an accurate head motion predictor to periodically inform where the user will be likely looking at, over a future horizon.

In this article, we consider the problem of predicting the user's head motion in 360\degree\ videos over a future horizon, based both and only on the past trajectory and on the video content.
Various methods tackling this problem with deep neural networks have therefore been proposed in the last couple of years (e.g., \cite{xu2018modeling,xu2018gaze,nguyen2018your,chinacom18,fan2017fixation}).
We show that the relevant existing methods have hidden flaws, that we thoroughly analyze to overcome with a new proposal establishing state-of-the-art performance.
We hence make two main contributions.

\noindent\textbf{Contributions}:\\
\noindent$\bullet$ \textbf{Uncovering hidden flaws of existing methods and performing a root-cause analysis}:\\
After a review and taxonomy of the most relevant and recent methods (PAMI18 \cite{xu2018modeling}, CVPR18 \cite{xu2018gaze}, MM18 \cite{nguyen2018your}, ChinaCom18 \cite{chinacom18} and NOSSDAV17 \cite{fan2017fixation}), we compare them to common baselines. 
First, comparing against the \textit{trivial-static baseline}, we obtain the intriguing result that they all perform worse, on their exact original settings, metrics and datasets.
Second, we show it is indeed possible to outperform the \textit{trivial-static baseline} (and hence the existing methods) by designing a stronger baseline, named the \textit{deep-position-only baseline}: it is an LSTM-based architecture considering only the positional information, while the existing methods are meant to benefit both from the history of past positions and knowledge of the video content.
From there, we carry out a thorough root-cause analysis to understand why the existing methods perform worse than baselines that do not consider the content information.
Looking into the metrics and the data, we show that: 
\new{(i) evaluating only on some specific pieces of trajectories or specific videos, where the content is proved useful, does not change the comparison results}, 
and that (ii) the content can indeed inform the head position prediction, but for prediction horizons longer than 2 to 3 sec.. All these existing methods consider shorter horizons.
Looking into the neural network architectures, we identify that: (iii) when the provided content features are the ground-truth saliency, the only architecture not degrading away from the baseline is the one with a Recurrent Neural Network (RNN) layer dedicated to the positional input, but (iv) when fed with saliency estimated from the content, the performance of this architecture degrades away from the \textit{deep-position-only baseline} again.
\\
\noindent$\bullet$ \textbf{Introducing a new deep neural architecture achieving state-of-the-art performance on all the datasets of compared methods and all prediction horizons (0-5 sec.)}:\\
To overcome this difficulty, we re-examine the requirements on how both modalities (past positions and video content) should be considered given the structure of the problem. We support our reasoning with the concept of Structural-RNN, modeling the dynamic head motion prediction problem as a spatio-temporal graph. We obtain a new deep neural architecture, that we name TRACK. 
\new{TRACK establishes state-of-the-art performance on all the prediction horizons 0-5 sec. and all the datasets of the existing competitors. In the 2-5 sec. horizon, TRACK outperforms the second-best method by up to 20\% in orthodromic distance error on focus-type videos, i.e., videos with low-entropy saliency maps.}

Owing to the critical results and perspective we raise on the state-of-the-art, and in our concern for reproducibility, the experimental setup and datasets of each assessed method and all our codes, are provided online (detailed and illustrated) at \cite{us_gitlab}. This reproducible framework has already obtained an ACM reproducibility badge \cite{us_ODS}, and allows to easily test any future approach.

Sec. \ref{sec:pb_taxo} formulates the exact prediction problem considered, and presents a taxonomy of the existing methods as well as a detailed description of each. Sec. \ref{sec:comp_baselines} evaluates these methods against two baselines it introduces, the \textit{trivial-static baseline} and the \textit{deep-position-only baseline}.
Sec. \ref{sec:analysis_data} presents the first part of the root-cause analysis by analyzing the data, introducing the \textit{saliency-only baseline}. Sec. \ref{sec:analysis_arch} completes the root-cause analysis by analyzing the architectural choices. Sec. \ref{sec:TRACK} presents our reasoning to obtain our new prediction method, TRACK, which establishes state-of-the-art performance. Sec. \ref{sec:discussion} gives perspective and connects our work to most recent critical re-examinations of deep learning-based approaches for other application domains. Sec. \ref{sec:conclu} concludes the article.

\section{Review and taxonomy of existing head prediction methods}\label{sec:pb_taxo}

This section reviews the existing methods relevant for the problem we consider. We start by formulating the exact problem: it consists, at each video playback time $t$, in predicting the future user's head positions between $t$ and $t+H$, as illustrated in Fig. \ref{fig:headMotionPrediction} and represented in Fig. \ref{fig:sketch_predwindow}, with the only knowledge of this user's past positions and the (entire) video content.
We therefore do not consider methods aiming to predict the entire user trajectory from the start based on the content and on the starting point as, e.g., targeted by the challenge in \cite{gutierrez2018introducing} or summarizing a 360$^\circ$ video into 2D \cite{Pano2vid,DeepRanking}. 
As well, and importantly, we consider that the users' statistics for the video are \textit{not} known at test time, hence we do not consider methods relying on these per-video statistics, such as \cite{Gwendal_clust,Laura_clust}.
Also, the domain of egocentric videos is related to that of 360$^\circ$ video. However, the assumptions are not exactly the same: only part of the scene and some regions likely to attract the users are available (video shot from a mobile phone), contrary to a 360$^\circ$ video. We therefore do not compare with such works.
The problem we tackle is inherently dynamic and aims to help streaming decisions made along the playback.
We then present the existing methods and classify them based on the choices of deep neural network architecture.
Finally, we provide a detailed description of each method we analyze later in this article.

\subsection{Problem formulation}\label{sec_probformul}
\begin{figure}[h]
  \centering
  \includegraphics[width=1.0\columnwidth]{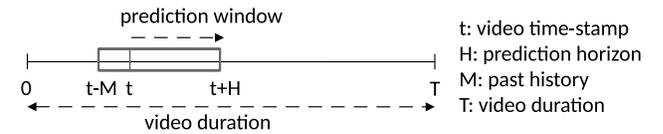}
  \caption{For each time-stamp $t$, all the next positions from $t$ until $t+H$ are predicted.}
  \label{fig:sketch_predwindow}
\end{figure}

Let us first define some notation.
Let $\mathbf{P}_t=[\theta_t,\varphi_t]$ denote the vector coordinates of the FoV at time $t$.
Let $\mathbf{V}_t$ denote the considered visual information at time $t$: depending on the models' assumptions, it can either be the raw frame with each RGB channel, or a 2D saliency map resulting from a pre-computed saliency extractor.
Let $T$ be the video duration.
The prediction is not assessed over the first $T_{start}$ seconds of video. To match the settings of the works we compare with, $T_{start}$ is set to $0$ sec. for all the curves generated in Sec. \ref{sec:comp_baselines}. In order to skip the exploration phase, as explained in Sec. \ref{sec:A2}, and be more favorable to all methods as they are not able to consider non-stationarity of the motion process, we set $T_{start}=6$ sec. from Sec. \ref{sec:analysis_data} onward.
We now refer to Fig. \ref{fig:sketch_predwindow}.
Let H be the \textit{prediction horizon}. 
We define the terms \textit{prediction step} $s$, and video \textit{time-stamp} $t$, such that:
at every \textit{time-stamp} $t\in[T_{start},T]$, we run predictions $\hat{\mathbf{P}}_{t+s}$, for all \textit{prediction steps} $s\in[0,H]$.

We formulate the problem of trajectory prediction as finding the best model $\mathbf{F}_H^*$ verifying:
\begin{align*}
\mathbf{F}_H^*=&\arg\min \mathbb{E}_t\Big[D\Big(\big[\mathbf{P}_{t+1},\dots,\mathbf{P}_{t+H}\big],\\
&\mathbf{F}_H\big(\left[\mathbf{P}_t,\mathbf{P}_{t-1},\dots,\mathbf{P}_0,\mathbf{V}_{t+1},\dots,\mathbf{V}_{t+H}\right]\big)\Big)\Big]
\end{align*}
where $D\left(\cdot\right)$ is the chosen distance between the ground-truth series of the future positions and the series of predicted positions. 
Except for the results in Fig. \ref{fig:difficulty_plots}, for each $s$, we average the errors $dist(\hat{\mathbf{P}}_{t+s},\mathbf{P}_{t+s})$ over all $t\in[T_{start},T]$.
As considered in the existing methods we compare with, we make $H$ vary between 0.2 sec. and 2.5 sec., then extend $H$ to 5 sec. as detailed from the analysis in Sec. \ref{sec:analysis_data}.

\subsection{Taxonomy}\label{sec:taxonomy}

\begin{figure}[h]
\centering
  \includegraphics[width=0.4\columnwidth]{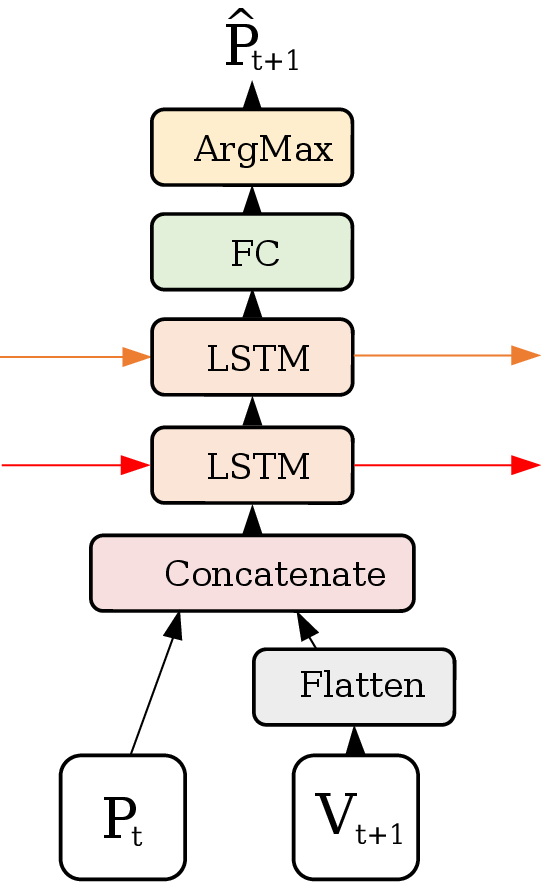}
  \hfill
  \includegraphics[width=0.4\columnwidth]{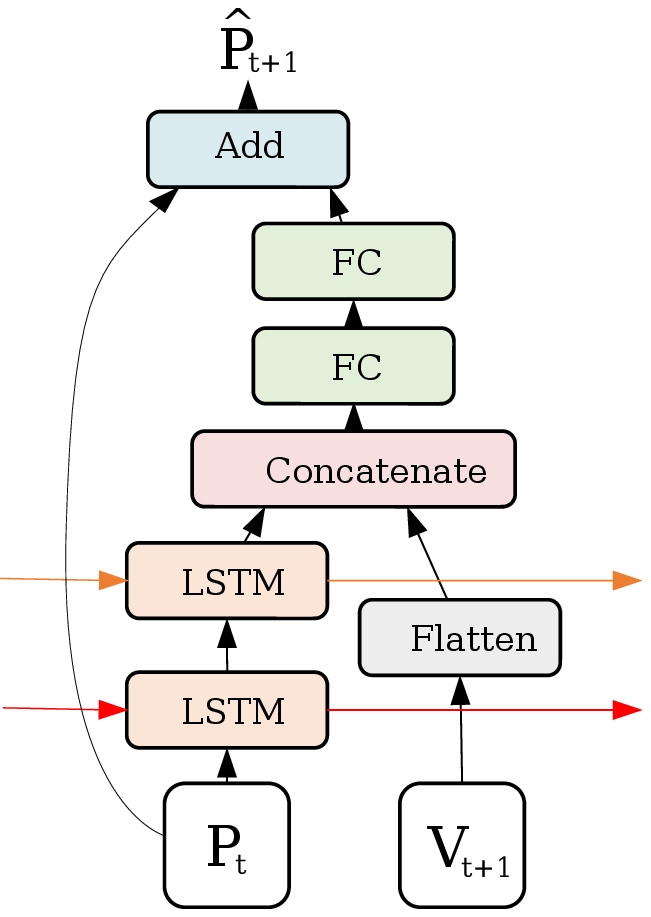}
  \caption{The building blocks in charge, at each time step, of processing positional information $P_t$ and content information $V_t$, that are visual features learned end-to-end or obtained from a saliency extractor module (omitted in this scheme). Left: MM18 \cite{nguyen2018your}. Right: CVPR18. \cite{xu2018gaze}}
  \label{fig:block_NOSSDAV_CVPR}
\end{figure}

Various approaches to predict user motion in 360\degree\ video environments have been published in the last couple of years, and are organized in Table \ref{table:taxonomy}. 
First, for the sake of clarity, each considered method is named with the name of the conference or journal it was published in, appended with the year of publication, as represented in column 1 of Table \ref{table:taxonomy} (starting from the left).
They consider different objectives (col. 2), such as predicting the future head position, gaze position or tiles in the FoV. The prediction horizons (col. 3) also span a wide range, from 30ms to 2.5 sec.. Some articles share common datasets for experiments (col. 4), while generally not comparing with each other. Different types of input and input formats are considered (col. 5): some consider the positional information implicitly by only processing the content in the FoV (PAMI18), other consider the position separately, represented as a series of coordinates (e.g., CVPR18) or as a mask (e.g., MM18), with the last sample only (IC3D17) or various length of history, some extract features from the visual content by employing some pre-trained saliency extractors (e.g. NOSSDAV17, MM18) or training end-to-end representation layers made of convolutional and max-pooling layers (e.g., PAMI18).
Finally, most of the methods but the first two in Table \ref{table:taxonomy} rely on deep-learning approaches.
A key aspect is the way they handle the combination of the positional information (if they consider it individually) with the video content information. As these two types of information are time series, those works all consider the use of deep Recurrent Neural Networks (RNN), and all use Long Short Term Memory (LSTM). However, whether the features are first extracted from each time series independently, or whether the time series samples are first concatenated then fed to a common LSTM, depends on each method. The positioning of the recurrent network in the whole architecture is the multimodality fusion criterion we have selected (col. 6) to order the rows in Table \ref{table:taxonomy} (within each group, methods are ordered from the most recently published), thereby extracting 3 groups of methods: 
\begin{itemize}
 \item if the positional information is not explicitly considered, then no combination is made and a single LSTM processes the content of the FoV: PAMI18;
 \item combination is made after the single LSTM module in CVPR18: the LSTM processes past positions, and its output gets fused with the video features through a fully connected layer (see Fig. \ref{fig:block_NOSSDAV_CVPR}-Right);
 \item if the current saliency map extracted from the content is first concatenated with the current position information, then the LSTM module handles both pieces of information modalities simultaneously: NOSSDAV17, ChinaCom18, MM18 (see Fig. \ref{fig:block_NOSSDAV_CVPR}-Left).
\end{itemize}

The architectures tackling this dynamic head motion prediction problem have hence three main objectives: (O1) extracting attention-driving features from the video content, (O2) processing the time series of position, and (O3) combining (fusing) both information modalities to produce the final position estimate. 
We depict the modules in charge of (O2) and (O3) of methods MM18 and CVPR18 in Fig. \ref{fig:block_NOSSDAV_CVPR}.
The existing methods are described more in detail next and those in bold are selected for comparison with the baselines presented in Sec. \ref{sec:comp_baselines}.


\begin{table*}
\begin{center}
{\small
\begin{tabularx}{\textwidth}{llllll}
\toprule
\parbox{0.10\textwidth}{Reference}
& \parbox{0.12\textwidth}{Objective}
& \parbox{0.10\textwidth}{Prediction\\horizon}
& \parbox{0.20\textwidth}{Dataset}
& \parbox{0.25\textwidth}{Inputs}
& \parbox{0.10\textwidth}{RNN \\before/after \\concatenation \\of modalities}\\
\midrule
\parbox{0.10\textwidth}{\textbf{PAMI18} \\\cite{xu2018modeling}} 
& \parbox{0.12\textwidth}{head \\coordinates} 
& \parbox{0.10\textwidth}{30ms} 
& \parbox{0.20\textwidth}{76 videos, \\58 users} 
& \parbox{0.25\textwidth}{frame cropped to FoV} 
& \parbox{0.10\textwidth}{N/A \\(no fusion)}\\
\parbox{0.10\textwidth}{IC3D17 \\\cite{IC3D17}} 
& \parbox{0.12\textwidth}{head \\coordinates} 
& \parbox{0.10\textwidth}{2s} 
& \parbox{0.20\textwidth}{16 videos, \\61 users} 
& \parbox{0.25\textwidth}{Pre-trained sal. in FoV} 
& \parbox{0.10\textwidth}{N/A \\(no fusion, \\no LSTM)}\\
\parbox{0.10\textwidth}{ICME18 \\\cite{ICME18}} 
& \parbox{0.12\textwidth}{tiles \\in FoV} 
& \parbox{0.10\textwidth}{6s} 
& \parbox{0.20\textwidth}{18 videos, \\48 users} 
& \parbox{0.25\textwidth}{Position history, \\users' distribution}
& \parbox{0.10\textwidth}{N/A \\(no LSTM)}\\
\cmidrule(ll){1-6}
\parbox{0.10\textwidth}{\textbf{CVPR18} \\\cite{xu2018gaze}} 
& \parbox{0.12\textwidth}{gaze \\coordinates} 
& \parbox{0.10\textwidth}{1s} 
& \parbox{0.20\textwidth}{208 videos, \\30+ users} 
& \parbox{0.25\textwidth}{Video frame, \\position history as coordinates}
& \parbox{0.10\textwidth}{before}\\
\cmidrule(ll){1-6}
\parbox{0.10\textwidth}{\textbf{MM18} \\\cite{nguyen2018your}}
& \parbox{0.12\textwidth}{tiles \\in FoV} 
& \parbox{0.10\textwidth}{2.5s} 
& \parbox{0.20\textwidth}{11 videos, \\48+ users from \cite{corbillondatasetMMSys, wutanwangyang} \\with custom pre-processing}
& \parbox{0.25\textwidth}{Pre-trained saliency, \\mask of positions} 
& \parbox{0.10\textwidth}{after}\\
\parbox{0.10\textwidth}{\textbf{ChinaCom18} \\\cite{chinacom18}} 
& \parbox{0.12\textwidth}{tiles \\in FoV} 
& \parbox{0.10\textwidth}{1s} 
& \parbox{0.20\textwidth}{NOSSDAV17's dataset} 
& \parbox{0.25\textwidth}{Pre-trained saliency, \\FoV tile history}
& \parbox{0.10\textwidth}{after}\\
\parbox{0.10\textwidth}{\textbf{NOSSDAV17} \\\cite{fan2017fixation}} 
& \parbox{0.12\textwidth}{tiles \\in FoV} 
& \parbox{0.10\textwidth}{1s} 
& \parbox{0.20\textwidth}{10 videos, \\25 users} 
& \parbox{0.25\textwidth}{Pre-trained saliency, \\FoV position or \\tile history}
& \parbox{0.10\textwidth}{after}\\
\bottomrule
\end{tabularx}
}
\end{center}
\caption{Taxonomy of existing dynamic head-prediction methods. References in bold are considered for comparison in Sec. \ref{sec:comp_baselines}.}
\label{table:taxonomy}
\end{table*}

\subsection{Detailed description of methods selected for comparison}

\noindent\textbf{PAMI18}: Xu et al. in \cite{xu2018modeling} design a Deep Reinforcement Learning model to predict head motion. Their deep neural network only receives the viewer's FoV as a $42\times42$ input image, and must decide to which direction and with which magnitude the viewer's head will move. Features obtained from convolutional layers processing each 360$^\circ$ frame cropped to the FoV are then fed into an LSTM to extract direction and magnitude. The training is done end-to-end. The prediction horizon is only one frame, i.e., 30ms. By only injecting the FoV, the authors make the choice not to consider the positional information explicitly as input. The PAMI18 architecture therefore does not feature any specific fusion module. The better performance of our \textit{deep-position-only baseline} shown in Sec. \ref{sec:comp_baselines} questions this choice.\\
\noindent\textbf{IC3D17}: The strategy presented by Aladagli et al. in \cite{IC3D17} simply extracts saliency from the current frame with an off-the-shelf method, identifies the most salient point, and predicts the next FoV to be centered on this most salient point. It then builds recursively. We therefore consider that this method to be a sub-case of PAMI18, and that the comparison with PAMI18 is thus more relevant.\\
\noindent\textbf{ICME18}: Ban et al. in \cite{ICME18} assume the knowledge of the users' statistics, and hence assume more information than our problem definition, which is to predict the user motion only based on the user's position history and the video content. We therefore do not consider this architecture for comparison. A linear regressor is first learned to get a first prediction of the displacement, which it then adjusts by computing the centroid of the k nearest neighbors corresponding to other users' positions at the next time-step.\\
\noindent\textbf{CVPR18}: In \cite{xu2018gaze}, Xu et al. predict the gaze positions over the next second in 360\degree\ videos based on the gaze coordinates in the past second and the video content.
As depicted in Fig. \ref{fig:block_NOSSDAV_CVPR}-Right, the time series of past head coordinates is processed by a doubly-stacked LSTMs. For the video information, spatial and temporal saliency maps are first concatenated with the RGB image, then fed to Inception-ResNet-V2 to obtain the ``saliency features'' denoted as $V{t+1}$ in Fig. \ref{fig:block_NOSSDAV_CVPR}-Right.
They formulate the gaze prediction problem the same way as the head prediction problem.\\
\noindent\textbf{MM18}: Nguyen et al. in \cite{nguyen2018your} first construct a saliency model based on a deep convolutional network and named PanoSalNet. 
The so-extracted saliency map is then fed, along with the position encoded as a mask, into a doubly-stacked LSTM, as shown in Fig. \ref{fig:block_NOSSDAV_CVPR}-Left.\\
\noindent\textbf{ChinaCom18}: Li et al. in \cite{chinacom18} present a similar approach as MM18, adding a correction module to compensate for the fact that tiles predicted to be in the FoV with highest probability may not correspond to the actual FoV shape (having even disconnected regions). This is a major drawback of the tile-based approaches as re-establishing FoV continuity may significantly impact final performance.\\
\noindent\textbf{NOSSDAV17}: Fan et al. in \cite{fan2017fixation} propose two LSTM-based networks, predicting the likelihood that tiles pertain to future FoV. Visual features extracted from a pre-trained VGG-16 network are concatenated with positional information, then fed into LSTM cells for the past $M$ time-steps, to predict the head orientations in the future $H$ time-steps.
Similarly to MM18 and as depicted in Fig. \ref{fig:block_NOSSDAV_CVPR}-Left, the building block of NOSSDAV17 first concatenates flattened saliency map and position, and feeds it to a doubly-stacked LSTM whose output is post-processed to produce the position estimate. An extended version of this work has been published in \cite{fan2019optimizing}.\\
These methods therefore make for a wide range of deep network architectural choices. In particular the fusion problem (O3), defined in Sec. \ref{sec:taxonomy}, may be handled differently. MM18 and CVPR18 are selected as representatives: combining both modalities before or after the recurrent (LSTM) unit, respectively.
There is no pairwise comparison between any of the above works.
From the above articles, the only works which provided their code and their deep neural networks for reproducibility are PAMI18 and MM18.
However, we could obtain all the datasets to compare with all (the datasets not publicly available were kindly shared by the authors whom we have contacted).

\subsection{Description of datasets from the literature for comparison}\label{sec:datasets}
This information is summarized in Table \ref{table:taxonomy}, and detailed here:\\
\noindent\textbf{PAMI18}: This dataset contains both head movement and eye movement data of 58 subjects on 76 360$^\circ$ videos of variable duration, from 10 to 80 seconds (25 seconds in average).\\
\noindent\textbf{CVPR18}: It is made of 208 360$^\circ$ videos between 15 and 80 seconds (36s in average), each video is viewed by at least 31 participants.\\
\noindent\textbf{NOSSDAV17}: This dataset consists of 10 360$^\circ$ videos with a duration of 60 seconds, along with the identification number of the tiles that overlap with the FoV of the viewer according to the head orientation data (the tile size considered is $192\times192$). This dataset contains the traces of 50 participants, however, for the experiment performed in \cite{fan2017fixation}, the traces of 25 participants only were used.\\
\noindent\textbf{MM18}: The dataset used in MM18 consists on the post-processing of two publicly available datasets \cite{corbillondatasetMMSys, wutanwangyang}. The first dataset \cite{wutanwangyang} includes 18 videos viewed by 48 users, from which 9 videos are selected. The second dataset \cite{corbillondatasetMMSys} has five videos viewed by 59 users, from which 2 videos are used. From the chosen videos, a segment is selected such that there are one or more events that introduce a new salient region (e.g. a scene change).\\
\noindent\textbf{MMSys18}: We also considered the dataset presented by David et al. in \cite{david2018dataset} and referred to as MMSys18. It is made of 19, 360$^\circ$ videos of 20 seconds, along with the head positions of 57 participants starting their exploration at a random angular position.

\section{Comparison against two baselines: trivial-static and deep-position-only}\label{sec:comp_baselines}

To compare the above recent proposals (PAMI18, CVPR18, MM18, ChinaCom18, NOSSDAV17) to a common reference, we first introduce the \textit{trivial-static baseline}. First, we show that all these methods on their original settings, metrics and datasets, are outperformed by this trivial baseline. This is surprising and raises the question of whether it is actually possible to learn anything meaningful with these settings (datasets and prediction horizons). To answer this question, we then introduce a \textit{deep-position-only baseline}, that we design as a sequence-to-sequence LSTM-based architecture exploiting the time series of past positions only (disregarding the video content). We show this new baseline is indeed able to outperform the \textit{trivial-static baseline} (establishing state-of-the-art performance). 
Later, Sec. \ref{sec:analysis_data} introduces a \textit{saliency-only baseline}.

\subsection{Definition of the trivial-static baseline}
Different linear predictors can be considered as baselines. We consider here the simplest one which predicts no motion: $\big[\hat{\mathbf{P}}_{t+1},\dots,\hat{\mathbf{P}}_{t+H}\big]=$$\big[\mathbf{P}_{t},\dots,\mathbf{P}_{t}\big]$. 

More complex baselines exist. For example in \cite{Baoshooting}, a Linear Regressor and a Neural Network perform better than the \textit{trivial-static baseline}. However, as we will see, all existing methods trying to leverage both the video content and the position to predict future positions perform worse than the \textit{trivial-static baseline}, without exception.

\subsection{Design of the deep-position-only baseline}

We now present an LSTM-based predictor which considers positional information only. 
An LSTM enables non-linear shape of the motion and the memory effect due to inertia, as discussed in \cite{xu2018gaze} and shown by the generated trajectories in \cite{us_gitlab}. 
We select a sequence-to-sequence (seq2seq) architecture because it has proven powerful at capturing complex dependencies and generating realistic sequences, as shown in text translation for which it has been introduced \cite{sutskever2014sequence}.

\begin{figure}[h]
  \centering
  \includegraphics[width=1.0\columnwidth]{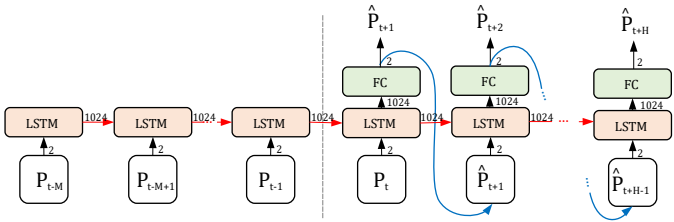}
  \caption{The \textit{deep-position-only baseline} based on an encoder-decoder (seq2seq) architecture.}
  \label{figure:fixationPrediction}
\end{figure}
As depicted in Fig.~\ref{figure:fixationPrediction}, a seq2seq framework consists of an encoder and a decoder. The encoder receives the \textit{historic window} input (samples from $t-M$ to $t-1$ shown in Fig. \ref{fig:sketch_predwindow}) and generates an internal representation. The decoder receives the output of the encoder and progressively produces predictions over the target horizon, by re-injecting the previous prediction as input for the new prediction time-steps.
This is a strong baseline (not only a trivial-static or linear predictor) processing the head coordinates only. 
We have optimized the \textit{deep-position-only baseline} as described in \cite{us_gitlab} and \cite[Sec. I]{us_supplemental}. This baseline has been inspired from the work of Martinez et al. in \cite{martinez2017human}, which re-examined major deep networks as multi-modal fusion methods, combining video content and motion time series for 3D human skeleton motion prediction. Their findings that all state-of-the-art methods were worse than a simple baseline, have echoed and corroborate with our own findings for the problem of multi-modal fusion methods for head motion prediction as detailed below. We give more perspective on this aspect in Sec. \ref{sec:discussion}.

\textbf{Reproducibility}: All the additional details of implementation are described in the supplemental material joined to the submission \cite[Sec. II]{us_supplemental}. We emphasize that the entire reproducible framework, with all methods including baselines, homogeneized datasets and common metrics, has been published in \cite{us_ODS}, obtained an ACM reproducibility badge, and is available in full at \cite{us_gitlab}.

\subsection{Results}\label{sec:results_literature}

We now present the comparisons of the state-of-the-art methods presented in Sec. \ref{sec:taxonomy} with the \textit{trivial-static baseline} and \textit{deep-position-only baseline} defined above.
We report the exact results of the original articles, along with the results of our baselines, the \textit{deep-position-only baseline} being trained and tested on the exact same train and test subsets of the original dataset as the original method (there is no training for the \textit{trivial-static baseline}). The benchmark metrics (related to predicting head or gaze positions, or FoV tiles) are those from the original articles, so are the considered prediction horizons $H$. 

\begin{figure}[ht!]
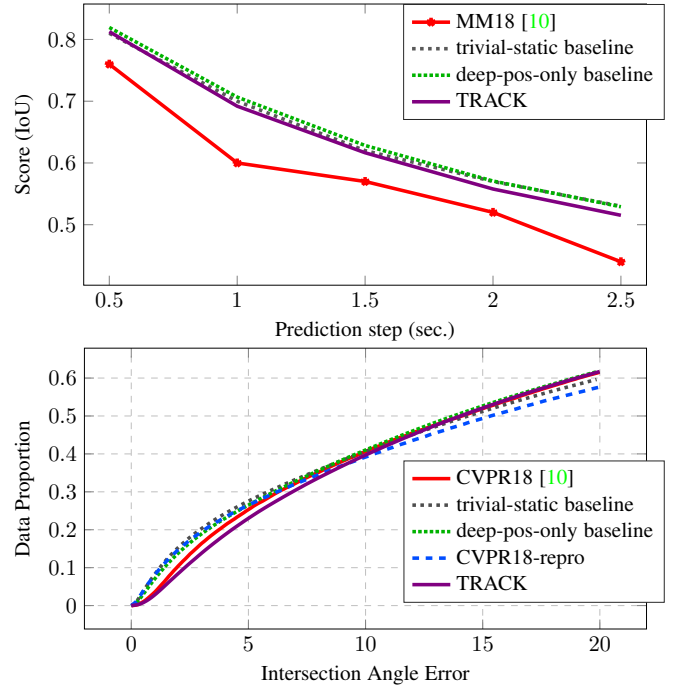

    \centering
    \input{Figures/MM18_Exper}\vfill
    \input{Figures/CVPR18_Exp}
    \caption{\textbf{Top}: Comparison with MM18 \cite{nguyen2018your}, $H=$2.5 seconds. \textbf{Bottom}: Comparison with CVPR18 \cite{xu2018gaze}, prediction horizon $H=1$ sec. CVPR18-repro is introduced in Sec. \ref{sec:metrics}, the model TRACK in Sec.~\ref{sec:with_CBsal}.}
    \label{fig:ComparisonCVPR18}
\end{figure}

Results for PAMI18 are shown in Table \ref{table:Res_PAMI18}, for CVPR18 in Fig. \ref{fig:ComparisonCVPR18}-\new{Bottom}, for MM18 in Fig. \ref{fig:ComparisonCVPR18}-\new{Top}, for ChinaCom18 in Table \ref{table:Res_ChinaCom18} and for NOSSDAV17 in Table \ref{table:Res_NOSSDAV17}.
Let us mention that none of these methods considered baselines identical to the \textit{trivial-static baseline} and \textit{deep-position-only baseline} defined above.
All perform worse than both our \textit{trivial-static} and \textit{deep-position-only baselines}. Specifically, all but one (CVPR18) perform significantly worse.
\begin{table*}[t]
\begin{center}
\begin{adjustbox}{width=1.0\textwidth}
\begin{tabular}{ccccccccccccccccc}
\toprule
Method & KingKong & SpaceWar2 & StarryPolar & Dancing & Guitar & BTSRun & InsideCar & RioOlympics & Average \\
\cmidrule(ll){1-10}
PAMI18~\cite{xu2018modeling} & 0.809 & 0.763 & 0.549 & 0.859 & 0.785 & 0.878 & 0.847 & 0.820 & 0.753 \\
\textit{trivial-static baseline} & 0.974 & 0.963 & 0.906 & 0.979 & 0.970 & 0.983 & 0.976 & 0.966 & 0.968 \\
\textit{deep-position-only baseline} & \textbf{0.983} & \textbf{0.977} & \textbf{0.930} & \textbf{0.984} & \textbf{0.977} & \textbf{0.987} & \textbf{0.982} & \textbf{0.976} & \textbf{0.977}\\
TRACK & 0.974 & 0.964 & 0.912 & 0.978 & 0.968 & 0.982 & 0.974 & 0.965 & 0.968 \\
\toprule
Method & SpaceWar & CMLauncher2 & Waterfall & Sunset & BlueWorld & Symphony & WaitingForLove & & Average \\
\cmidrule(ll){1-10}
PAMI18~\cite{xu2018modeling} & 0.626 & 0.763 & 0.667 & 0.659 & 0.693 & 0.747 & 0.863 & & 0.753 \\
\textit{trivial-static baseline} & 0.965 & 0.981 & 0.973 & 0.964 & 0.970 & 0.968 & 0.978 & & 0.968 \\
\textit{deep-position-only baseline} & \textbf{0.976} & \textbf{0.989} & \textbf{0.984} & \textbf{0.973} & \textbf{0.979} & \textbf{0.976} & \textbf{0.982} & & \textbf{0.977}\\
TRACK & 0.965 & 0.981 & 0.972 & 0.964 & 0.970 & 0.969 & 0.977 & & 0.968 \\
\bottomrule
\end{tabular}
\end{adjustbox}
\end{center}
\caption{Comparison with PAMI18 \cite{xu2018modeling}: Mean Overlap scores of FoV prediction, prediction horizon $H\approx 30ms$ (1 frame). The model TRACK is introduced in Sec.~\ref{sec:with_CBsal}.}
\label{table:Res_PAMI18}
\end{table*}

\begin{table}
\begin{center}
\begin{adjustbox}{width=1.0\columnwidth}
{\small
\begin{tabular}{lcc|cc|cc}
\toprule
& \multicolumn{2}{l}{\textit{trivial-static baseline}} & \multicolumn{2}{l}{\textit{deep-position-only baseline}} & \multicolumn{2}{l}{ChinaCom18} \\
& Accuracy & F-score & Accuracy & F-score & Accuracy & F-score \\
\midrule
Hog Rider & 96.29\% & 0.8858 & \textbf{96.97\%} & \textbf{0.9066} & 77.09\% & 0.2742 \\
Driving with & 95.96\% & 0.8750 & \textbf{96.59}\% & \textbf{0.9843} & 77.34\% & 0.2821 \\
Shark Shipwreck & 95.23\% & 0.8727 & \textbf{96.12\%} & \textbf{0.8965} & 83.26\% & 0.5259 \\
Mega Coaster & 97.20\% & 0.9144 & \textbf{97.71\%} & \textbf{0.9299} & 88.90\% & 0.7011 \\
Roller Coaster & 96.99\% & 0.9104 & \textbf{97.50\%} & \textbf{0.9256} & 88.28\% & 0.6693 \\
Chariot-Race & 97.07\% & 0.8802 & \textbf{96.91\%} & \textbf{0.9056} & 87.79\% & 0.6040 \\
SFR Sport & 96.00\% & 0.8772 & \textbf{96.91\%} & \textbf{0.9054} & 89.29\% & 0.7282 \\
Pac-Man & 96.83\% & 0.8985 & \textbf{97.16\%} & \textbf{0.9089} & 87.45\% & 0.6826 \\
Peris Panel & 95.60\% & 0.8661 & \textbf{96.54\%} & \textbf{0.8947} & 89.12\% & 0.7246 \\
Kangaroo Island & 95.35\% & 0.8593 & \textbf{96.54\%} & \textbf{0.8954} & 82.62\% & 0.5308 \\
\midrule
Average & 96.15\% & 0.8840 & \textbf{96.90\%} & \textbf{0.9063} & 72.54\% & 0.5155 \\
\bottomrule
\end{tabular}
}
\end{adjustbox}
\end{center}
\caption{Comparison with ChinaCom18 \cite{chinacom18}, prediction horizon $H=1$ second.}
\label{table:Res_ChinaCom18}
\end{table}

\begin{table}
\begin{center}
\begin{adjustbox}{width=1.0\columnwidth}
{\small
\begin{tabular}{lccc}
\toprule
Method & Accuracy & F-Score & Rank Loss \\
\midrule
NOSSDAV17-Tile~\cite{fan2017fixation} & 84.22\% & 0.53 & 0.19 \\
NOSSDAV17-Orient.~\cite{fan2017fixation} & 86.35\% & 0.62 & 0.14 \\
\textit{trivial-static baseline} & 95.79\% & 0.87 & 0.10 \\
\textit{deep-position-only baseline} & \textbf{96.30}\% & \textbf{0.89} & \textbf{0.09}\\
TRACK & 95.48\% & 0.85 & 0.15\\
\bottomrule
\end{tabular}
}
\end{adjustbox}
\end{center}
\caption{Comparison with NOSSDAV17: Performance of Tile- and Orientation-based networks of \cite{fan2017fixation} compared against our \textit{deep-position-only baseline}, prediction horizon $H=1$ second. The model TRACK is introduced in Sec.~\ref{sec:with_CBsal}.}
\label{table:Res_NOSSDAV17}
\end{table}

We define below the metrics used for every considered predictor:\\
\noindent$\bullet$ \textbf{NOSSDAV17} \cite{fan2017fixation} considers the following metrics:\\
$-$ \textit{Accuracy}: ratio of correctly classified tiles to the union of predicted and viewed tiles.\\
$-$ \textit{Ranking Loss}: number of tile pairs that are incorrectly ordered by probability normalized to the number of tiles.\\
$-$ \textit{F-Score}: harmonic mean of \textit{precision} and \textit{recall}, where \textit{precision} is the ratio of correctly predicted tiles by the total number of predicted tiles, and \textit{recall} is the ratio of correctly predicted tiles by the number of viewed tiles.\\
Let us point out here that the tile data is not balanced, as more tiles pertain to class 0 (tile $\not\in$ FoV) than to class 1 (tile $\in$ FoV) owing to the restricted size of the FoV compared to the complete panoramic size. If we predict all the tiles systematically in class 0, the accuracy already gets to 83.86\%. 
The accuracy is indeed known to be a weak metric to measure the performance of such unbalanced datasets.\\
\noindent$\bullet$ \textbf{PAMI18} \cite{xu2018modeling} uses as metric the Mean Overlap (MO) defined as:
\begin{align*}
  MO = \cfrac{A(FoV_{p} \cap FoV_{g})}{A(FoV_{p} \cup FoV_{g})}
\label{equation:meanoverlap}
\end{align*}
Where $FoV_{p}$ is the predicted FoV, $FoV_{g}$ is the ground-truth FoV, and $A(\cdot)$ is the area of a panoramic region.\\
\noindent$\bullet$ \textbf{CVPR18} \cite{xu2018gaze} uses the Intersection Angle Error $IAE$ for each gaze point $(\theta, \varphi)$ and its prediction $(\hat{\theta}, \hat{\varphi})$, defined as $IAE = arccos(\langle P,\hat{P}  \rangle)$, where $P$ is the 3D coordinate in the unit sphere:\\
$P = (x, y, z) = (cos(\theta)cos(\varphi), cos(\theta)sin(\varphi), sin(\theta))$. 
Let us mention that CVPR18 also considers a deep-position-only baseline. However, ours appears stronger, likely due to the seq2seq architecture. We readily apply our different predictors on the gaze data available in the CVPR18-dataset.
\\
\noindent$\bullet$ \textbf{MM18} \cite{nguyen2018your} takes the tile with the highest viewing probability as the center of the predicted viewport, and assigns it and all the neighboring tiles that cover the viewport, with label 1. Tiles outside the viewport are assigned 0. Then, the score is computed on these labels as $IoU = TP/TT$ (True Positive $TP$, True Total $TT$), the intersection between prediction and ground-truth of tiles with label 1 ($TP$) over the union of all tiles with label 1 in the prediction and in the ground-truth ($TT$).\\
\noindent$\bullet$ \textbf{ChinaCom18} \cite{chinacom18} uses the Accuracy and F-Score on the labels assigned to each predicted tile.
%


\section{Root cause analysis: the metrics in question}\label{sec:analysis_metrics}
We have shown that the existing methods assessed above, which try to leverage both positional information and video content to predict future positions, perform worse than a simple baseline assuming no motion, which in turn can be outperformed by the \textit{deep-position-only baseline} (considering only positional information).
This section and the next two (Sec. \ref{sec:analysis_data} \& Sec. \ref{sec:analysis_arch}) aim to identify the reasons why the existing approaches perform worse than the baselines. In this part, we focus on the possible causes due to the evaluation, specifically asking:\\

\noindent\textit{\textbf{Q1 Metrics: \new{Can the methods perform better than the baselines for some specific videos or pieces of trajectories?}}}

\subsection{Evaluation Metrics}\label{sec:metrics}
Let us first describe the losses and evaluation metrics considered from now on. The prediction of the FoV motion can be cast as a classification problem, where pixels or tiles are classified in or out of future FoV (as done in NOSSDAV17, MM18, ChinaCom18). However, this problem is inherently imbalanced. Therefore, for the analysis, we choose to keep the original formulation as a regression problem.
The tracking problem on a sphere can be assessed by different distances.
Given two points on the surface of the unit sphere $P_{1}=(\theta_{1}, \varphi_{1})$ and $P_{2}=(\theta_{2}, \varphi_{2})$, where $\theta$ is the longitude and $\varphi$ is the latitude of the point, possible distances are:\\
\noindent$\bullet$ Mean squared error $= ((\theta_1-\theta_2)^2+(\varphi_1-\varphi_2)^2)/2$\\
\noindent$\bullet$ Angular error $=\sqrt{\arctan(\sin(\Delta\theta)/\cos(\Delta\theta))^2+(\varphi_1-\varphi_2)^2}$, where $\Delta\theta = \theta_1-\theta_2$\\
\noindent$\bullet$ Orthodromic distance\\ $=\arccos{(\cos{(\varphi_1)}\cos{(\varphi_2)}\cos{(\Delta\theta)} + \sin{(\varphi_1)}\sin{(\varphi_2)})}$ 
which is a reformulation of:
\begin{equation}
\label{eq:orthodromic_distance}
D(P_1, P_2) = \arccos{(\vec{P_1} \bullet \vec{P_2})},
\end{equation}
where $\bullet$ is the dot product operation, and $\vec{P_1}$ are the coordinates in the unit sphere of point $P$. Indeed, for a point $P_1=(\theta_{1}, \varphi_{1})$, the coordinates in the unit sphere are then given by $\vec{P_1} = (\cos{\theta_{1}}\cos{\varphi_{1}}, \sin{\theta_{1}}\cos{\varphi_{1}}, \sin{\varphi_{1}})$.\\
The latter two metrics are able to handle the periodicity of the latitude, which the first one cannot. The difference between \textit{angular error} and \textit{orthodromic distance} is that the latter computes the distance on the surface of the sphere, while the \textit{angular error} computes the error of each angle independently. Finally, owing to its adequacy to the tracking problem on the unit sphere, we choose the \textbf{orthodromic distance} as the test metric in our approach.

\subsection{Q1: \new{Can the methods perform better than the baselines for some specific pieces of trajectories or videos?}}
\new{
The metrics used in Sec. 3 are averages over time trajectories and videos. The question we ask is whether the methods can perform better than the baselines for some specific pieces of trajectories or videos.}

\subsubsection{Specific pieces of trajectory}
\begin{figure}[!ht]
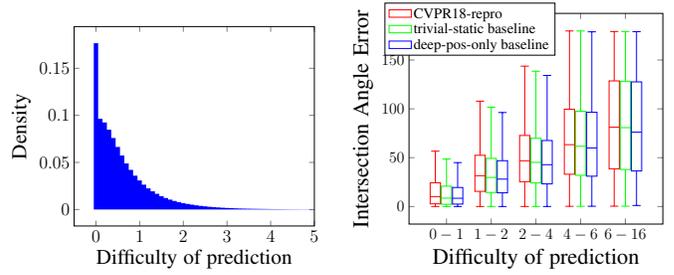

\centering
\input{Figures/2nd_derivative_CVPR18}\hfill
\input{Figures/BoxplotsDifficulty}
\caption{Left: Distribution of difficulty in the CVPR18 dataset. Right: Error as a function of the difficulty for the CVPR18-repro model.}
\label{fig:difficulty_plots}
\end{figure}

\new{
To evaluate whether the existing methods perform better than the baselines in some specific pieces of the trajectory, we adopt the same approach as in \cite[Sec. 4]{Alahi_CVPR2016}, introducing the \textit{Average non-linear displacement error} as a metric to evaluate the error around the non-linear regions of the trajectory where most errors occur owing to human-content interactions.
We therefore quantify the \textit{difficulty of prediction} with the second derivative of the trajectory, i.e., the radius of curvature.}
To obtain detailed results (for each instant of time of each user and video pair), we re-implement CVPR18 with the exact same architectural and training parameters as those described in the article \cite{xu2018gaze}.\footnote{We had to replicate the architecture of CVPR18 because we could not find any official code and the authors did not reply to our emails. Our reproduced code is available online at \cite{us_gitlab} and detailed in \cite[Sec. III]{us_supplemental}.}
\new{
The curve CVPR18-repro in Fig. \ref{fig:ComparisonCVPR18}-Bottom shows that we obtain similar results on the original dataset (higher on the first half of the truncated CDF, then slightly lower on the second half of the truncated CDF). This confirms the validity of our re-implementation.
Fig. \ref{fig:difficulty_plots}-Left depicts the distribution of the prediction difficulty. Fig. \ref{fig:difficulty_plots}-Right shows that for every difficulty range, CVPR18-repro is not able to improve the prediction over the baselines.
Considering CVPR18 and MM18 the two representative and best performing methods in Sec. 3 (apart from the baselines), for the sake of space we also report the results for MM18 in the supplemental material in \cite[Sec. IV]{us_supplemental}. We obtained similar qualitative results with MM18.}

\new{We conclude that for more difficult parts of the trajectory, the CVPR18-repro or MM18 methods are not able to improve over the baselines.}

\subsubsection{Specific videos}
\new{
Fig. \ref{fig:difficulty_plots}-Left shows that the majority of the data is in the 0-1 difficulty range, therefore, we can think the models have difficulty to pay attention to the rarer cases of trajectory pieces where the prediction difficulty is higher. 
To evaluate whether the existing methods perform better than the baselines when the dataset (train and test sets) is properly balanced with videos where the content is proved to help, we consider the dataset prepared in Sec. \ref{sec:with_GTsal}.
The details on the usefulness of the content are given in Sec. \ref{sec:analysis_data}. The performance of CVPR18-repro and MM18-repro on this dataset can be seen}
\new{in Fig. \ref{fig:orig_improved} in average and per test video in \cite[Sec. V]{us_supplemental}: they are never able to take advantage of the content as they are systematically outperformed by the \textit{deep-position-only baseline} (even for the videos where the saliency is proved useful). \\
}

\noindent\textbf{Answer to Q1}: \new{No, the methods considering the video content do not perform better than the \textit{deep-position-only baseline} for specific pieces of trajectories or videos where the knowledge of the content should improve the prediction.}


\section{Root cause analysis: the data in question}\label{sec:analysis_data}
In this section, we focus on the possible causes due to the data. In Sec. \ref{sec:analysis_arch}, we analyze the possible architectural causes.
This section therefore aim to answer question Q2, whose answer is provided at the end of the section:\\

\noindent\textit{\textbf{Q2 Data: Do the datasets (made of videos and motion traces) match the design assumptions the methods build on?}}\\
To answer Q2, we consider the assumptions at the core of the existing architectures attempting to leverage the knowledge of position history and video content, and break them down into :
\begin{itemize}
    \item Assumption (A1): \new{the position history can inform the prediction of future positions}
    \item Assumption (A2): \new{the visual content can inform the prediction of future positions}
\end{itemize}
We identify whether these assumptions hold in the datasets and settings considered by the existing methods (Sec. \ref{sec:A1} and \ref{sec:A2}), and introduce a new \textit{saliency-only baseline} to do so (Sec. \ref{sec:sal-baseline}).

\subsection{Assumption (A1): \new{the position history is informative of future positions}}\label{sec:A1}
\begin{figure}
\centering
\input{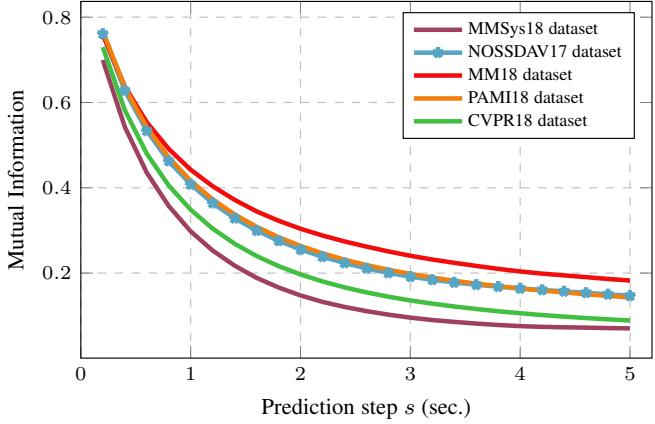}
\caption{\new{Mutual information $I(P_t;P_{t+s})$ between $P_t$ and $P_{t+s}$ (averaged over $t$ and videos) for all the datasets used in NOSSDAV17, PAMI18, CVPR18 and MM18, with the addition of MMSys18.}}
\label{fig:MI}
\end{figure}

\new{The amount of information held by a process about another one can be quantified by the Mutual Information (MI). This in turns informs on the degree of predictability of the target process using the first process. MI has been used in \cite{Rossi_MMVE} for inter-user analysis. Here, we define the MI between head positions of a given user at time $t$ and $t+s$ by
$I(P_t;P_{t+s}) = D_{KL}(Pr[P_t,P_{t+s}]||Pr[P_t]\otimes Pr[P_{t+s}])$, where $D_{KL}(\cdot)$ and $\otimes$ stand for the Kullback–Leibler divergence and convolutional product, respectively.
For each of the datasets considered in PAMI18, CVPR18, MM18, NOSSDAV17, and MMSys18, Fig. \ref{fig:MI} represents MI normalized and averaged over all videos and time stamp $t$, as a function of $s\in[0,H=5sec.]$. The 2D-coordinates have been discretized in 128 bins.}
\new{This figure shows that position at time $t+s$ can be predicted to a significant degree by $P_t$ when $s$ is low (e.g., lower than 2 sec.). As expected, the further away the prediction step, the lowest the predictability of $P_{t+s}$ from $P_t$.}
\new{In \cite[Sec. VI]{us_supplemental}, we also relate MI with a more intuitive characterization of the datasets, showing that the amount of user's motion is generally low, except in the MMSys18 dataset.} 

\noindent\textit{Does Assumption (A1) hold?}: \new{On the datasets and prediction horizons considered in the literature ($H\leq$ 2 sec.), the position history is therefore strongly informative of the next positions.} Another element supporting this observation is the best performance obtained by our baseline exploiting position only (see Sec. \ref{sec:comp_baselines} above).
A similar study was conducted in \cite{Baoshooting} showing that the viewer motion has a strong temporal auto-correlation. 

\subsection{Definition of the saliency-only baseline}\label{sec:sal-baseline}

To analyze Assumption A2 in Sec. \ref{sec:A2} and assess how much gain can the consideration of the content bring to the prediction, we first define a so-called \textit{saliency-only baseline}.
This baseline is defined from an attentional heat map, either extracted from the visual content (heat map then named Content-Based saliency) or directly from the position data of all the users (heat map then named Ground-Truth saliency).
For either type of heat map, the \textit{saliency-only baseline} provides an upper-bound on the prediction error that a more refined processing of the heat map, in combination with the past positions, would make.
In this section, we only consider heat maps obtained from the users data, we therefore start by defining such heat maps. Only in Sec. \ref{sec:with_CBsal} do we use the heat maps estimated from the video content.

\subsubsection{Definition of the ground-truth saliency}\label{sec:GTsal}
To be independent from the imperfection of any saliency predictor fed with the visual content, we consider here the ground-truth saliency: it is the heat map (2D distribution) of the viewing patterns, obtained at each point in time from the users' traces.
To compute the ground-truth saliency maps, we consider the point at the center of the viewport $P_{u,v}^{t}$ for user $u \in U$ and video $v \in V$ at time-stamp $t\in[0,T]$, where $T$ is the length of the trace. For each head position $P_{u,v}^{t}$, we compute the orthodromic distance $D(\cdot)$ from $P_{u,v}^{t}$ to each point $Q_{x,y}$ at longitude $x$ and latitude $y$ in the equirectangular frame. Then, we use a modification of the radial basis function (RBF) kernel shown in Eq.~\ref{eq:RBF kernel} to convolve the points in the equirectangular frame and obtain the Ground-Truth Saliency (${GT\_Sal}$) for user $u$ on video $v$ at time $t$ in image location $(x,y)$:

\begin{equation}
\label{eq:RBF kernel}
{GT\_Sal}_{u, v, x, y}^{t} = \exp{\left(-\frac{D(P_{u,v}^{t}, Q_{x,y})^2}{2\sigma^2}\right)},
\end{equation}
where $D(P_{u,v}^{t}, Q_{x,y})$ is the orthodromic distance, computed using Eq.~\ref{eq:orthodromic_distance}.
A value of $\sigma=6$\degree\ is chosen so that the ground-truth saliency maps look qualitatively similar to those of PanoSalNet \cite{nguyen2018your} used in Sec. \ref{sec:with_CBsal}.
We compute saliency maps ${GT\_Sal}_{u, v}^{t}$ per user $u \in U$, video $v \in V$ and time-stamp $t$ by convolving each head position $P_{u,v}^{t}$ with the modified RBF function in Eq~\ref{eq:RBF kernel}. The saliency map at time $t$ of video $v$ is calculated as $GT\_Sal_{v}^{t} = \frac{1}{U} \sum_{u \in U} {GT\_Sal}_{u, v}^{t}$, where $U$ is the total number of users watching this video. 

\subsubsection{Definition of the K-saliency-only baseline}
We extract the $K$ highest peaks of the heat map for every prediction step $t+s$ (for all $t$, for all $s\in[0,H]$). At every $t+s$, the \textit{K-saliency-only baseline} predicts $\hat{\mathbf{P}}_{t+s}$ as the position of the peak, amongst the $K$ peaks, which is closest to the last known user's position $\mathbf{P}_t$.

\begin{figure}[ht!]
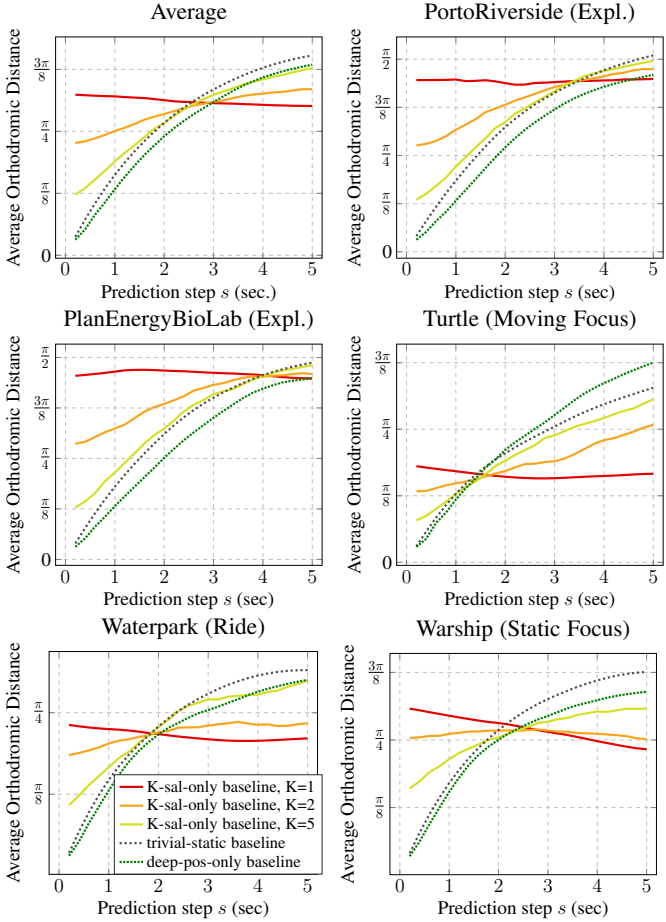

\input{Figures/Seq2SeqPerf_Average}\hfill
\input{Figures/Seq2SeqPerf_PortoRiverside}\vfill
\input{Figures/Seq2SeqPerf_PlanEnergyBioLab}\hfill
\input{Figures/Seq2SeqPerf_Turtle}\vfill
\input{Figures/Seq2SeqPerf_Waterpark}\hfill
\input{Figures/Seq2SeqPerf_Warship}
\caption{Prediction error on the MMSys18 dataset. The \textit{deep-position-only baseline} is tested on the 5 videos above, and trained on the others (see \cite[Sec. I]{us_supplemental} or \cite{us_ODS,us_gitlab}). Top left: Average results on all 5 test videos. Rest: Detailed result per video category (Exploration, Moving Focus, Ride, Static Focus). Legend is identical in all sub-figures.}
\label{fig:A2_MMSys-dataset}
\end{figure}

\begin{figure}[ht!]
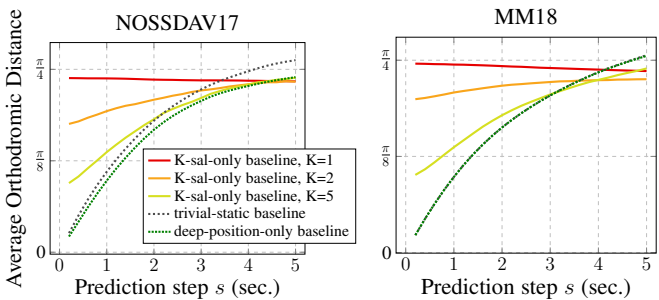

\input{Figures/Seq2SeqPerf_Average_Fan_NOSSDAV_17}\hfill
\input{Figures/Seq2SeqPerf_Average_Nguyen_MM_18}\hfill
\caption{Prediction error averaged on test videos of the datasets of NOSSDAV17 (left) and MM18 (right). We refer to the supplemental material \cite[Sec. II]{us_supplemental} or \cite{us_gitlab} for the train-test video split used for the \textit{deep-position-only baseline} (identical to original methods). Legend is identical in both sub-figures.}
\label{fig:A2_NOSSDAV-MM-dataset}
\end{figure}

Fig. \ref{fig:A2_MMSys-dataset} and \ref{fig:A2_NOSSDAV-MM-dataset} show the prediction error of the \textit{K-saliency-only baseline} for $K=1$, $2$, $5$. For low $s$, we verify that the higher $K$, the lower the error close to time $t$, because the more the number of points of interest possibly considered.
However, as the prediction step $s$ increases and $t+s$ gets away from $t$, the error is lower for lower $K$. Indeed, if the user moves, then she is more likely to get closer to a more popular point of interest, that is to a higher-ranked peak.

\subsubsection{Definition of the saliency-only baseline}\label{sec:salonlybaseline}
As mentioned in the beginning of the section, each \textbf{\textit{K-saliency-only baseline}} can be considered as an upper-bound on the error that the best predictor optimally combining position and content modality could get.
Therefore, for a given $\kappa$, we define the \textit{saliency-only baseline} as the minimum of these \textit{K-saliency-only baseline}, for $K\in[1,\kappa]$ and for every $s$ in $[0,H]$. In this article, we set $\kappa=5$. The \textit{saliency-only baseline} is shown in red in Fig. \ref{fig:A2_CVPR18-dataset}. From Fig. \ref{fig:A2_CVPR18-dataset}, we do not represent the \textit{K-saliency-only baselines} anymore, but only the \textit{saliency-only baseline}.

\subsection{Background on human attention in VR}
Before analyzing Assumption (A2), let us first provide some characteristics of the human attention in VR identified recently.
It has been recently shown in \cite{sitzmann2018saliency} and \cite{almquist2018prefetch} that, when presented with a new VR scene (the term ``scene'' is defined by Magliano and Zacks in \cite{magliano2011impact} as a period of the video between two edits with space discontinuity), a human first goes through an exploratory phase that lasts for about 10 to 15 sec. (\cite[Fig. 18]{almquist2018prefetch}, \cite[Fig. 2]{sitzmann2018saliency}), before settling down on so-called Regions of Interest (RoIs), that are salient areas of the content. The duration and amplitude of exploration, as well as the intensity of RoI fixation, depend on the video content itself. Almquist et al.~\cite{almquist2018prefetch} have identified the following main video categories for which they could discriminate significantly different users' behaviors: \emph{Exploration, Static focus, Moving focus} and \emph{Rides}. In \emph{Exploration} videos, the spatial distribution of the users' head positions tends to be more widespread, making harder to predict where the users will watch and possibly focus on. 
\emph{Static focus} videos are made of a single salient object (e.g., a standing-still person), making the task of predicting where the user will watch easier in the focus phase. In \emph{Moving focus} videos, contrary to \emph{Static focus} videos, the RoIs move over the sphere and hence the angular sector where the FoV will be likely positioned changes over time.
\emph{Rides} videos are characterized by substantial camera motion, the attracting angular sector being likely that of the direction of the camera motion.

\subsection{Assumption (A2): the visual content is informative of future positions}\label{sec:A2}
\begin{figure}
\centering
\input{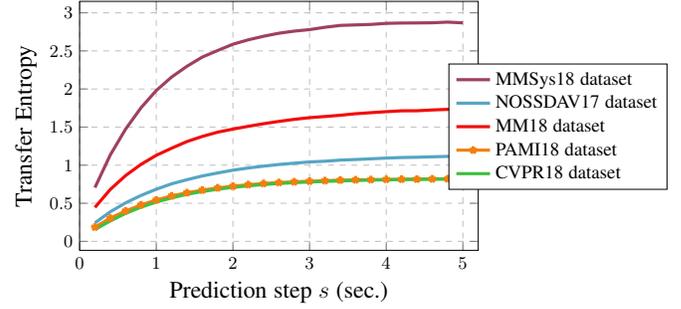}
\caption{\new{Transfer Entropy (TE) $TE_{V\rightarrow P}(t,s)$ between $V_{t+s}$ and $P_{t+s}$ (averaged over $t$ and videos) for all the datasets used in NOSSDAV17, PAMI18, CVPR18 and MM18, with the addition of MMSys18.}}
\label{fig:CB_TE}
\end{figure}

\begin{figure}[ht!]
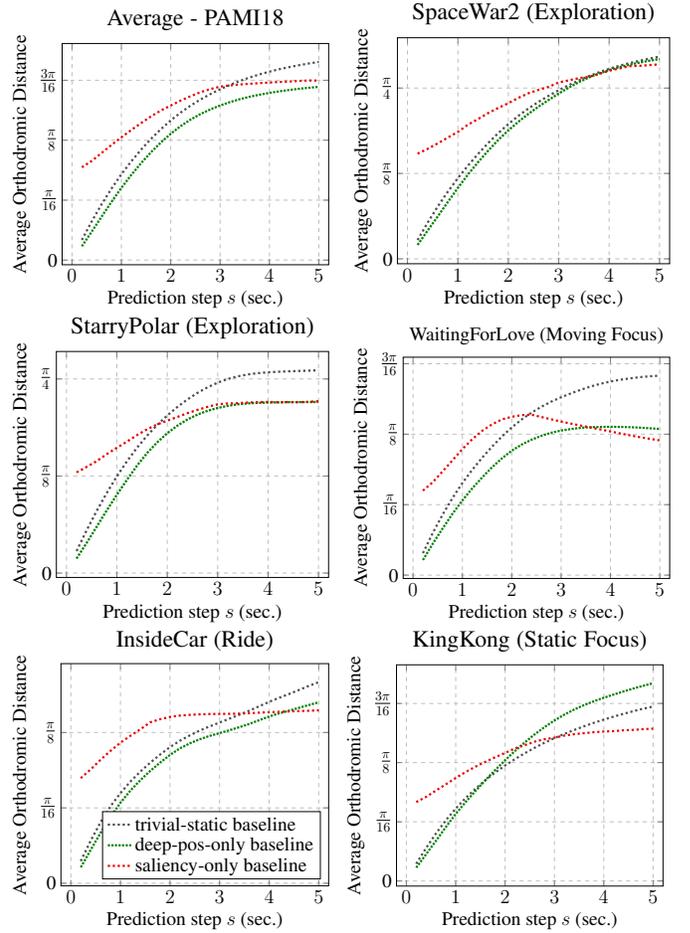

\input{Figures/Seq2SeqPerf_Average_Xu_PAMI_18}\hfill
\input{Figures/Seq2SeqPerf_Average_Xu_PAMI_18_Expl_Spacewar2}\vfill
\input{Figures/Seq2SeqPerf_Average_Xu_PAMI_18_Expl_StarryPolar}\hfill
\input{Figures/Seq2SeqPerf_Average_Xu_PAMI_18_Moving}\vfill
\input{Figures/Seq2SeqPerf_Average_Xu_PAMI_18_Rides}\hfill
\input{Figures/Seq2SeqPerf_Average_Xu_PAMI_18_Static}
\caption{Prediction error on the dataset of PAMI18. We refer to the supplemental material \cite[Sec. II]{us_supplemental} or \cite{us_gitlab} for the train-test video split used for the \textit{deep-position-only baseline} (identical to original method). Top left: Average on test videos. Rest: Results per video category (Exploration, Moving Focus, Ride, Static Focus). Legend is identical in all sub-figures.}
\label{fig:A2_PAMI18-dataset}
\end{figure}

\begin{figure}[ht!]
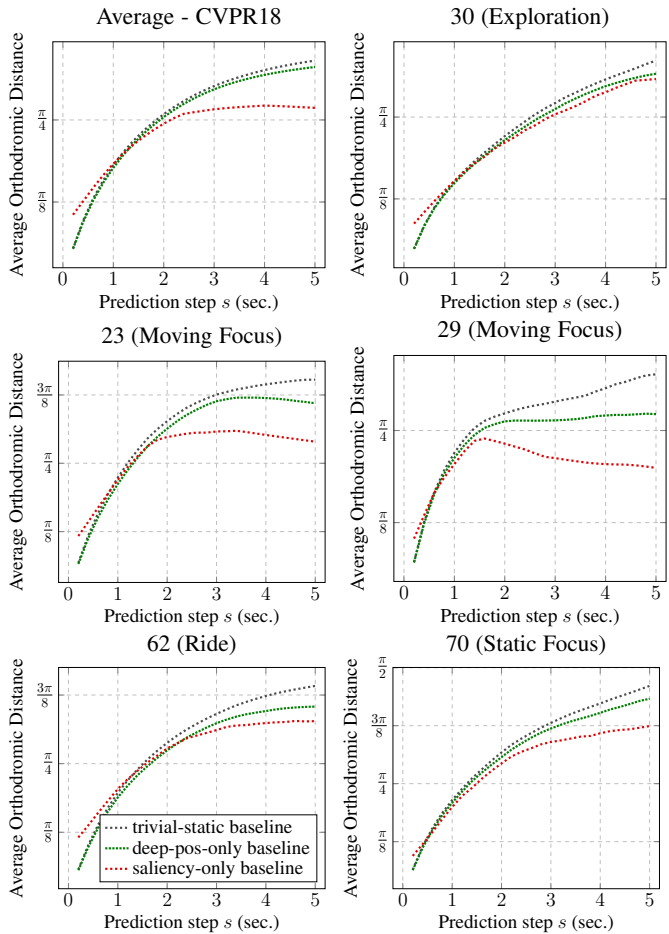

\input{Figures/Seq2SeqPerf_Average_Xu_CVPR_18}\hfill
\input{Figures/Seq2SeqPerf_Average_Xu_CVPR_18_Explor}\vfill
\input{Figures/Seq2SeqPerf_Average_Xu_CVPR_18_Moving_23}\hfill
\input{Figures/Seq2SeqPerf_Average_Xu_CVPR_18_Moving}\vfill
\input{Figures/Seq2SeqPerf_Average_Xu_CVPR_18_Rides}\hfill
\input{Figures/Seq2SeqPerf_Average_Xu_CVPR_18_Static}
\caption{Prediction error on the dataset of CVPR18. We refer to the supplemental material \cite[Sec. II]{us_supplemental} or \cite{us_gitlab} for the train-test video split used for the \textit{deep-position-only baseline} (identical to original method). Top left: Average on test videos. Rest: Results per video category (Exploration, Moving Focus, Ride, Static Focus). Legend is identical in all sub-figures.}
\label{fig:A2_CVPR18-dataset}
\end{figure}

We now analyze whether this assumption (A2) holds, and for which settings (datasets, prediction horizons).
\new{As for (A1), we first quantify how much additional information can be gained on $P_{t+s}$ by knowing the visual content $V_{t+s}$ at time $t+s$, given we already know the past positions. This corresponds to the conditional MI $I(P_{t+s};V_{t+s}|P_t)$, also named Transfer Entropy (TE) and satisfying for every video:
$TE_{V\rightarrow P}(t,s) = I(P_{t+s};V_{t+s}|P_t) = H(P_{t+s}|P_t) - H(P_{t+s}|P_t,V_{t+s})$, where $H(\cdot)$ denotes the entropy.
TE has been used in \cite{Rossi_MMVE} but not with saliency data. Fig. \ref{fig:CB_TE} represents $TE_{V\rightarrow P}(t,s)$ averaged over all time stamps $t$ and videos of every dataset. The 2D-coordinates have been discretized in 128 bins and $V_{t+s}$ is taken as the Content-Based saliency defined in Sec. \ref{sec:with_CBsal}, the probability values being discretized} into 256 bins.
\new{The TE values cannot be compared across the datasets, but the important observation is that the TE value triples from $s=0$ to $s=5$ sec. It shows that the predictability of future positions from the content, conditioned on the position history, is initially low then increases with $s$.}
The results of MI in Fig. \ref{fig:MI} and TE in Fig. \ref{fig:CB_TE} therefore show that short-term motion is mostly driven by inertia from $t$, while the content saliency may impact the trajectory in the longer-term. To cover both short-term and long-term, we set the prediction horizon $H=5$ sec.. We confirm this and better quantify the durations of both phases for the different video categories in the next results.
We analyze A2 on the datasets used in NOSSDAV17, MM18, CVPR18 and PAMI18. We also consider an additional dataset, referred to as MMsys18-dataset \cite{david2018dataset}. All these datasets are detailed in Sec. \ref{sec:datasets}.
In MMsys18-dataset, the authors show that the exploration phase in their videos lasts between 5 and 10s, and show that after this initial period, the different users' positions have a correlation coefficient reaching 0.4 \cite[Fig. 4]{david2018dataset}. This dataset is made of 12 Exploration videos, 4 Static focus videos (Gazafisherman, Sofa, Mattswift, Warship), 1 Moving focus video (Turtle) and 2 Ride videos (Waterpark and Cockpit).
Fig. \ref{fig:A2_MMSys-dataset}, \ref{fig:A2_NOSSDAV-MM-dataset}, \ref{fig:A2_CVPR18-dataset} and \ref{fig:A2_PAMI18-dataset} depict the prediction error for prediction steps $s\in[0,H=5$ sec.$]$, obtained with the \textit{deep-position-only baseline} and \textit{saliency-only baseline} on the 4 previous datasets. We remind that each point for every given step $s$, is an average over all the users and all time-stamp $t\in[T_{start},T]$, with $T$ the video duration and $T_{start}=6$ sec. from now on to skip the initial exploration phase (presented right above in the beginning of this Sec. \ref{sec:A2}) and ensure that the content can be useful for all time-stamps $t$.
By analyzing the \textit{saliency-only baseline} for every prediction step $s$ (saliency baseline in red in Fig. \ref{fig:A2_CVPR18-dataset}), the same phenomenon can be observed on all the datasets: the \textit{saliency-only baseline} has a higher error than the \textit{deep-position-only baseline} for prediction steps $s$ lower than 2 to 3 seconds. This means that there is no guarantee that the prediction error over the first 2 to 3 seconds can be lowered by considering the content. After 2 to 3 sec., on non-Exploration videos, we can see that relevant information can be exploited from the heat maps to lower the prediction error compared to the \textit{deep-position-only baseline}. 
When we isolate the results per video type, e.g., in Fig. \ref{fig:A2_MMSys-dataset}, for Exploration (PortoRiverside, PlanEnergyBioLab), a Ride (WaterPark) a Static focus (Warship) and a Moving focus (Turtle) videos, we observe that the saliency information can significantly help predict the position for prediction steps beyond 2 to 3 seconds.

We therefore conclude by answering \\
\noindent\textit{\textbf{Q2 Data: Do the datasets (made of videos and motion traces) match the design assumptions the methods build on?}}\\

\subsection{Q2: Do the datasets (made of videos and motion traces) match the design assumptions the methods build on?}\label{sec:answerA2}
\noindent\textbf{Answer to Q2}:\\
\noindent$\bullet$ 
\new{Study of MI for assumption A1 confirms that the level of predictability of short-term position from past position is significant, corresponding to the inertia effect and frequent low velocity in some datasets.}\\
\noindent$\bullet$ Considering the ground-truth saliency (attentional heat maps), we conclude on A2 by stating that considering the content in the prediction can significantly help for non-Exploration videos if the prediction horizon is longer that 2 to 3 sec.. There is no guarantee it can significantly or easily help for shorter horizons.
All the selected existing works considered prediction horizons lower than 2.5 sec., making it very unlikely to outperform the \textit{deep-position-only baseline}.

Having shown it is difficult to outperform the \textit{deep-position-only baseline} on these short horizons, next we investigate why most existing methods are however not able to match its performance.

\section{Root cause analysis: the architectures in question}\label{sec:analysis_arch}

In Sec. \ref{sec:analysis_data}, we have analyzed the possible causes for the weakness of the existing predictors, related to the metrics and the assumptions on the dataset. As they do not suffice to explain the counter-performance of the existing methods compared with single-modality baselines, in this section, we state and analyze the possible architectural causes.
Let us recall the three main objectives a prediction architecture must meet, as stated in Sec. \ref{sec:taxonomy}: (O1) extracting attention-driving features from the video content, (O2) processing the time series of position, and (O3) fusing both information modalities to produce the final series of position estimates. 
Note that this is a conceptual description, and does not necessarily correspond to a processing sequence: fusion (O3) can be performed from the start and O1 and O2 may not be performed in distinguishable steps or elements, as it is the case in NOSSDAV17 or MM18.

\textbf{The main interrogation is}: Why does the performance (of existing predictors compared with baselines) degrade when both modalities are considered? To explore this question from the architectural point of view, we divide this in two intermediate questions \textit{Q3} and \textit{Q4}.\\

\noindent\textit{\textbf{Q3 on ground-truth saliency: If O1 is solved perfectly by providing the ground-truth saliency, what are good choices for O2 and O3?}}\\
That is, in comparison with the baselines considering each modality individually, choices whose performance improves, or at least does not degrade, when considering both information modalities.

\subsection{Answer to Q3 - Analysis with ground-truth saliency}\label{sec:with_GTsal}
In our taxonomy in Sec. \ref{sec:taxonomy}, we have distinguished the prediction methods that consider both input modalities, based on the way they handle the fusion: either both position and visual information are fed to a single RNN, in charge of at least O3 and O2 at the same time (case of MM18, ChinaCom18, NOSSDAV17), or the time series of positions are first processed with a dedicated RNN, the output of which then gets fused with visual features (case of CVPR18). To answer Q3, we consider their most recent representatives: the building blocks of MM18 and CVPR18 (see Fig. \ref{fig:block_NOSSDAV_CVPR}). We still consider that O1 is solved perfectly by considering the ground-truth saliency introduced in Sec. \ref{sec:A2}.

\noindent\textbf{Prediction horizon}: From the answer to Q2, we consider the problem of predicting head positions over a prediction horizon longer than the existing methods (see Table \ref{table:taxonomy}), namely 0 to $H=5$ seconds. This way, both short-term where the motion is mostly driven by inertia at $t$, and long-term where the content saliency impacts the trajectory, are covered.

\noindent\textbf{Dataset}: Given the properties of MMSys18-dataset, where users move significantly more (see Sec. \ref{sec:A1}) and which comprises different video categories (introduced in Sec. \ref{sec:A2}), we select this dataset for the next experiments investigating the architectures.
In particular, we draw a new dataset out of MMSys18-dataset, selecting 10 train and 4 test videos by making sure that the sets are balanced between videos where the content is helpful (Static focus, Moving focus and Rides) and those where it is not (Exploration). Specifically, the train set is made with 7 Exploration videos, 2 Static Focus and 1 Ride, while the test set has 2 Exploration, 1 Static focus and 1 Ride videos. This number of videos is equivalent to the dataset considered in MM18, ChinaCom18 and NOSSDAV17 (10).
This dataset is therefore challenging but also well fitted to assess prediction methods aiming to get the best out of positional and content information.

\noindent\textbf{Auto-regressive framework}:
Our re-implementation of CVPR18, named CVPR18-repro, has been introduced in Sec. \ref{sec:metrics}.
For MM18, we use the code provided by the authors in \cite{panosalnetRepo}. 
The evaluation metric is still the orthodromic distance as exposed in Sec. \ref{sec:A2}.
We make three modifications to CVPR18 and MM18 (shown in Fig. \ref{fig:block_NOSSDAV_CVPR}), which we refer to as CVPR18-improved and MM18-improved, respectively.
\textbf{First}, as for our \textit{deep-position-only baseline}, we add a sequence-to-sequence auto-regressive framework to predict over longer prediction windows. We therefore embed each MM18 and CVPR18 building blocks into the sequence-to-sequence framework.
It corresponds to replacing every LSTM cell in Fig. \ref{figure:fixationPrediction} with the building blocks represented in Fig. \ref{fig:block_NOSSDAV_CVPR}.
\textbf{Second}, we train them with the mean squared error based on 3D Euclidean coordinates $(x,y,z)\in\mathbb{R}^3$. This helps the convergence with a seq2seq framework handling content, which is likely due to the removal of the discontinuity of having to use a modulo after each output in the training stage when Euler angles are considered. With 3D Euclidean coordinates, the projection back onto the unit sphere is made only at test time. We however retain the orthodromic distance as the benchmark metric.
\textbf{Third}, instead of predicting the absolute position as done by MM18, we predict the displacement (motion). This corresponds to having a residual connection, which helps to reduce the error in the short-term, as also identified by \cite{martinez2017human}.
Specifically for the MM18 block, we also change (1) the saliency map that we grow from $16\times 9$ to $256\times 256$, and (2) the output, i.e. the center of the FoV, which is defined by its $(x,y,z)$ Euclidean coordinates.

\noindent\textbf{Training}:
We train each model for 500 epochs, with a batch size of 128, with Adam optimization algorithm with a learning rate of $0.0005$ and the mean squared error based on 3D Euclidean coordinates $(x,y,z)\in\mathbb{R}^3$ as loss function.

\begin{figure}[!t]
\centering
\input{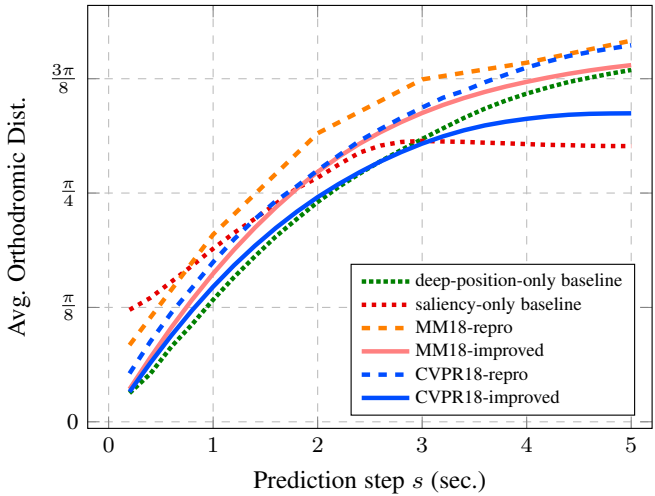}
\caption{Average prediction error of the original and improved models of MM18 and CVPR18, all fed with GT-sal, compared with baselines.}
\label{fig:orig_improved}
\end{figure}

\begin{figure}[t!]
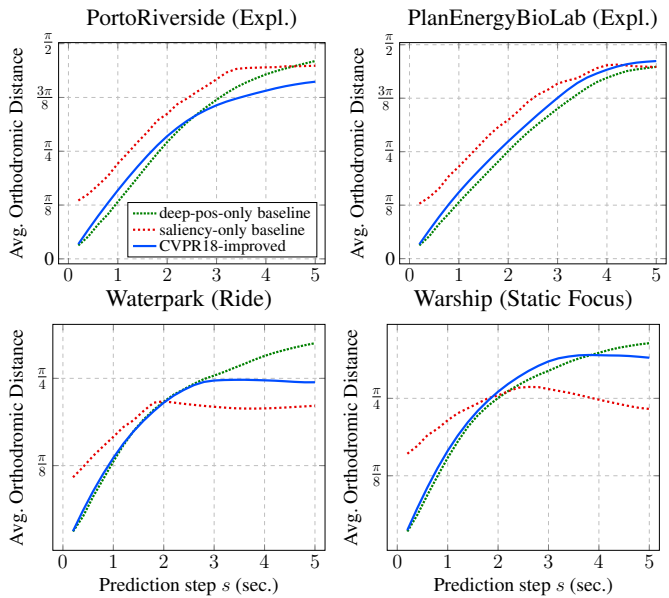

\centering
\input{Figures/CVPR18_Improv_Perf_PortoRiverside}\hfill
\input{Figures/CVPR18_Improv_Perf_PlanEnergyBioLab}\vfill
\input{Figures/CVPR18_Improv_Perf_Waterpark}\hfill
\input{Figures/CVPR18_Improv_Perf_Warship}\hfill
\caption{
Prediction error for CVPR18-improved. Detailed result for each type of test video. Legend is identical in all sub-figures.
}
\label{fig:improved_detailed}
\end{figure}

\noindent\textbf{Results}:
Fig. \ref{fig:orig_improved} shows \new{the improved models of MM18 and CVPR18 perform better than the original models. It also shows} that MM18-improved is still not able to perform at least as well as the \textit{deep-position-only baseline}. 
However, it is noticeable that CVPR18-improved is able to outperform the \textit{deep-position-only baseline} for long-term prediction, approaching the \textit{saliency-only baseline}. CVPR18-improved is also able to stick to the same performance as the \textit{deep-position-only baseline} for short-term prediction.
Fig. \ref{fig:improved_detailed} provides the detailed results of CVPR18-improved over the different videos in the test set, associated with their respective category identified in \cite{almquist2018prefetch}. While the average results show reasonable improvement towards the \textit{saliency-only baseline}, we observe that CVPR18-improved significantly improves over the \textit{deep-position-only baseline} for non-exploratory videos. Finally, we recall that the visual features provided to CVPR18-improved are the ground-truth saliency (i.e., the heat maps obtained from the users traces).

\noindent\textbf{Answer to Q3}: 
If O1 is solved perfectly by providing the ground-truth saliency, then O2 and O3 are best achieved separately by having a dedicated recurrent unit to extract features from the positional information only, before merging them in subsequent layers with visual features, as CVPR18 does. If the same recurrent unit is both in charge of O2 and O3, as in MM18, it appears to prevent from reaching the performance of the \textit{deep-position-only baseline}.

Therefore, we next analyze:\\

\noindent\textit{\textbf{Q4 on content-based saliency: If O1 is solved approximately by providing a saliency estimate obtained from the video content only, do the good choices for Q3 still hold, or does the performance degrade away from the baselines again? If so, how to correct?}}

\subsection{Answer to Q4 - Analysis with content-based saliency}\label{sec:with_CBsal}

We first summarize the findings of the root-cause analysis so far. In Q1, we found that even though averaging the prediction error over the trajectory might benefit the baselines, it does not and it is not a cause for the worse performance. In Q2, we have shown that the design assumption of the predictors are met if the dataset is made of non-exploratory videos with sufficient motion, and the prediction horizon is greater than 2 to 3 sec.. In Q3, on horizons and datasets verifying the latter conditions, we have found that when the visual information is represented by ground-truth saliency (O1 is perfectly solved), only the architecture of CVPR18 is able to exploit this modality without degrading compared with the baselines.

In this section, we do not consider O1 perfectly solved anymore. We consider the saliency information (i.e., heat map) is estimated from the video content only, not obtained from the users' statistics anymore.
Our goal is not to find the best saliency extractor for O1, but instead to uncover the impact of less accurate saliency information onto the architecture's performance, to then overcome this impact if necessary.

In the remainder of the paper, when the heat map fed to a method is obtained from the video content (not from the users traces), the name of the method is prefixed with CB-sal (for Content-Based saliency). Also, CB \textit{saliency-only baseline} denotes the \textit{saliency-only baseline} defined in Sec. \ref{sec:salonlybaseline} when the heat map is obtained from the content, and not from the users traces.
Conversely, when the heat map fed to a method is obtained from the users traces (and not estimated from the video content), the name of the method is prefixed with GT-sal (for Ground-Truth saliency, defined in Sec. \ref{sec:GTsal}). The GT \textit{saliency-only baseline} denotes the \textit{saliency-only baseline} defined in Sec. \ref{sec:salonlybaseline} when the heat map is obtained from the users traces.

\noindent\textbf{Saliency extractor}:
We consider PanoSalNet \cite{panosalnetRepo,nguyen2018your}, also considered in MM18. 
The architecture of PanoSalNet is composed by nine convolution layers, the first three layers are initialized with the parameters of VGG16 \cite{simonyan2014very}, the following layers are first trained on SALICON\cite{huang2015salicon}, and finally the entire model is re-trained on 400 pairs of video frames and saliency maps in equirectangular projection.
We exemplify the resulting saliency on a frame in \cite[Fig. 6]{us_ODS}.

\noindent\textbf{Results of CVPR18-improved}:
\begin{figure}[ht!]
\centering
\input{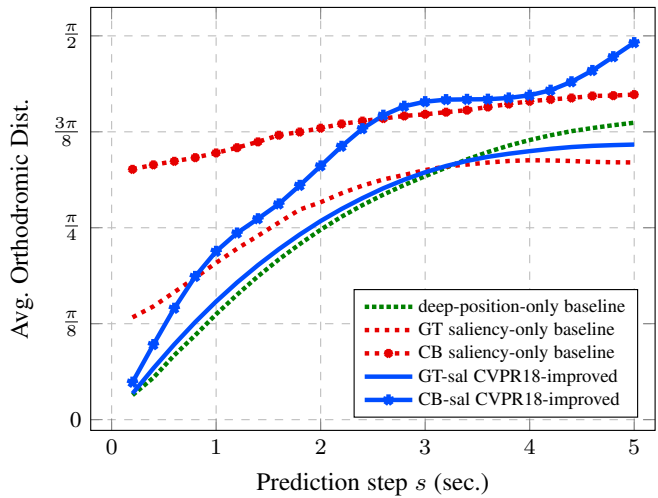}
\caption{Prediction error of CB-sal CVPR18-improved (with Content-Based saliency) against GT-sal CVPR18-improved (with Ground-Truth saliency) and baselines.}
\label{fig:content-saliency-avg}
\end{figure}
First, Fig. \ref{fig:content-saliency-avg} shows the expected degradation using the content-based saliency (obtained from PanoSalNet) compared with the ground-truth saliency:
the CB \textit{saliency-only baseline} (dashed red line) is much less accurate than the GT \textit{saliency-only baseline} (solid red line).\\
Second, we observe that, despite performing well with ground-truth saliency, CVPR18-improved fed with content-based saliency degrades again away from the \textit{deep-position-only baseline}.
Specifically, two questions arise:
\begin{itemize}
    \item Why does CB-sal CVPR18-improved degrades from GT-sal CVPR18-improved for horizons $H\leq 2$ sec., where the best to achieve is the \textit{deep-position-only baseline} according to Fig. \ref{fig:orig_improved}?\\
    The training losses are the same. The only difference is in the input values representing the saliency. We can show that the saliency CB-sal is less sparse than GT-sal, hence there are more non-zero inputs, which are also less accurate (obviously, compared to the GT). Therefore, the contribution of the CB-sal inputs should be nullified by the weights of the fully-connected layer in charge of the fusion. It is simple to verify that when fully connected layers have to cancel out part of their inputs acting as noise for the classification task, the convergence of the training error degrades with the number of such inputs. Such wrong performance in training indicates a sub-optimal architecture for the problem at hand.
    \item Why does CB-sal CVPR18-improved degrade from original CVPR18 for $H\in[0s,1 sec.]$?\\
    The first difference is the training loss, defined over a longer horizon for CB-sal CVPR18-improved ($H\in[0$ sec.,5 sec.$]$), while it is only for $H=1$ sec. in original CVPR18. The former loss is therefore likely more difficult to explore and minimize.
    The second difference is the presence, in original CPVR18, of convolutional and pooling layers processing various visual inputs including saliency before the fusion. Such layers can help decrease the input level into the fusion layer. However, they are not sufficient to enable a fully-connected layers to predict over $[0s,Hs]$ for $H\geq 3$ sec., as discussed in the next section.
\end{itemize}

\noindent\textbf{Partial answer to Q4}:
If O1 is solved approximately by providing a saliency estimate obtained from the video content only, the good choice for Q3 (CVPR18-improved) is not sufficient anymore.\\

\section{TRACK: A new architecture for content-based saliency}\label{sec:TRACK}

We now first complete the root-cause analysis by examining more detailedly the architectural reasons for CVPR18-improved to degrade away again from the baselines with CB-sal.
We then propose our new deep architecture, TRACK, stemming from this analysis. Its evaluation shows superior (once equal) performance on all the datasets of considered competitors and wider prediction horizons.

\subsection{Analysis of the problem with CVPR18-improved and content-based saliency (CB-sal)}\label{sec:analysis5}
The fundamental characteristic of the problem at hand is: over the prediction horizon, the relative importance of both modalities (past positions and content) varies.
Indeed, we expect the motion inertia to be more prominent first, and only then the content to possibly attract attention and change the course of the motion.
It is therefore crucial to have a way of combining both modality features in a time-dependent manner to produce the final prediction.
However, in the best-performing architecture so far, CVPR18-improved, we notice that the single RNN component enables this time-dependent modulation only for the positional features, while the importance of the content cannot be modulated over time.
Replacing the ground-truth saliency with content-based saliency, the saliency map becomes much less correlated with the positions to predict. It is therefore important to be able to attenuate its effect in the first prediction steps, and give it more importance in the later prediction step.

\subsection{Designing TRACK}
\begin{figure}[h]
  \centering
  \includegraphics[width=0.6\columnwidth]{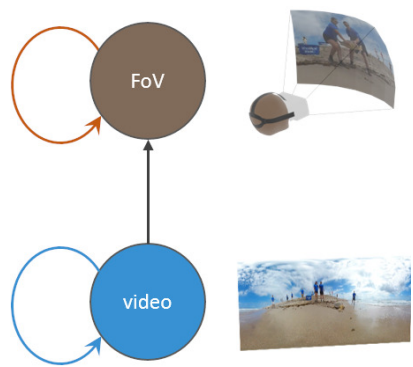}
  \caption{The dynamic head motion prediction problem cast as a spatio-temporal graph. Two specific edgeRNN corresponds to the brown (inertia) and blue (content) loops, a nodeRNN for the FoV encodes the fusion of both to result into the FoV position.}
  \label{fig:STgraph}
\end{figure}

\begin{figure*}[h]
  \centering
  \includegraphics[width=1.0\textwidth]{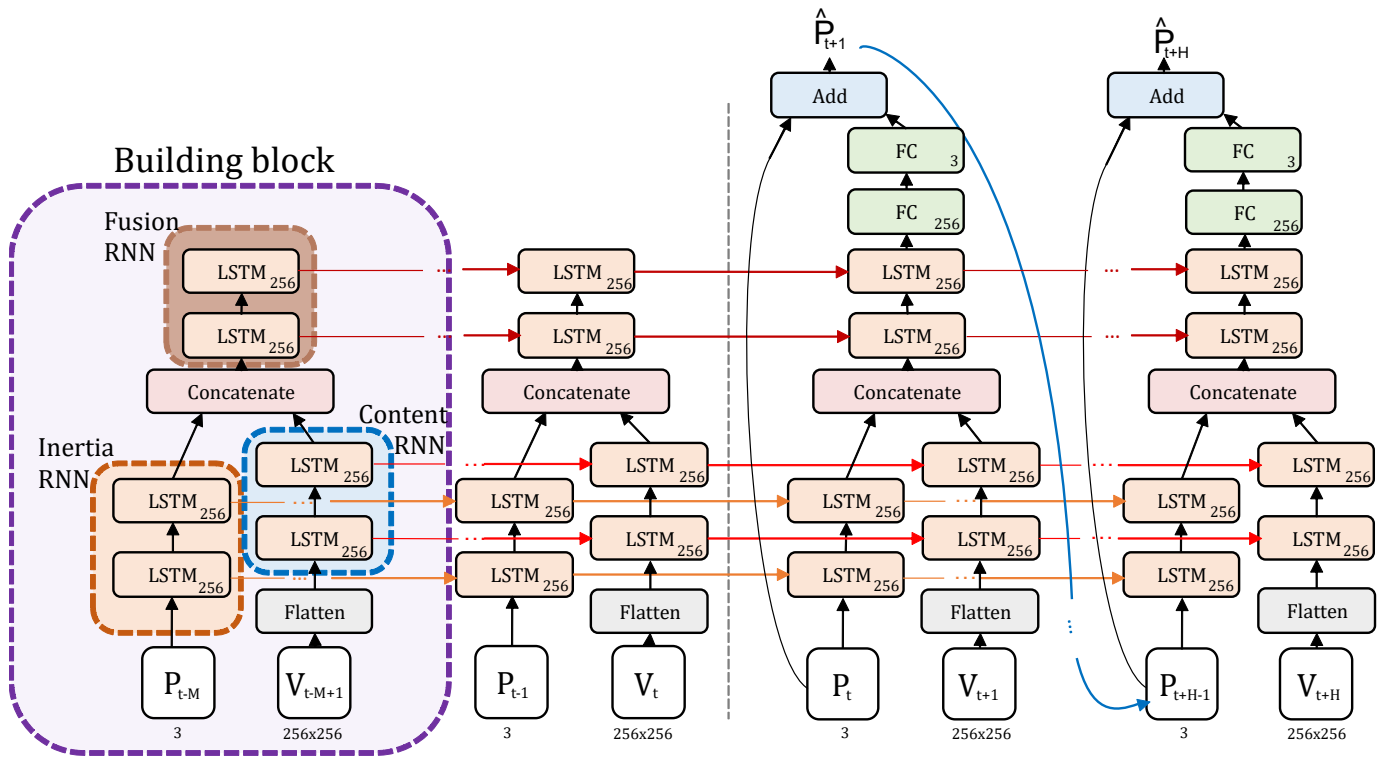}
  \caption{The proposed TRACK architecture. The colors refer to the components in Fig. \ref{fig:STgraph}: the building block (in purple) is made of a an Inertia RNN processing the previous position (light brown), a Content RNN processing the content-based saliency (blue) and a Fusion RNN merging both modalities (dark brown).}
  \label{fig:TRACK}
\end{figure*}

From the latter analysis, \textbf{a key architectural element to add is a RNN processing the visual features (such as CB-sal), before combining it with the positional features.}
Furthermore, this analysis connects with the seminal work of Jain et al., introducing Structural-RNN in \cite{jain2016structural}. It consists in casting a spatio-temporal graph describing a problem's structure into a rich RNN mixture following well-defined steps. 
Though the connection with head motion prediction is not direct, we can formulate our problem structure in the same terms.
First, two contributing factor components are involved: the user's FoV and the video content.
We can therefore express the spatio-temporal graph of a human watching a 360\degree\ video in a headset as shown in Fig. \ref{fig:STgraph}.
Second, these two components are semantically different, and are therefore associated with: (i) an edgeRNN and a nodeRNN for the FoV, (ii) an edgeRNN for the video (only one input to the node), resulting in the architectural block shown in purple in Fig. \ref{fig:TRACK}. Embedded into a sequence-to-sequence framework, we name this architecture \textbf{TRACK}.

\noindent\textbf{Components of TRACK}: The modules of TRACK are represented by (i) a doubly-stacked LSTM with 256 units each, processing the flattened CB-saliency map pre-generated for each time-stamp; (ii) another set of doubly-stacked LSTM with 256 units to process the head orientation input; (iii) a third set of doubly-stacked LSTM with 256 units to handle the multimodal fusion; and finally (iv) a FC layer with 256 and a FC layer with 3 neurons is used to predict the (x,y,z) coordinates, as described in Sec. \ref{sec:analysis_arch}.

\subsection{Evaluation of TRACK}

\subsubsection{Comparison with GT-sal CVPR18-improved}

\begin{figure}
\centering
\input{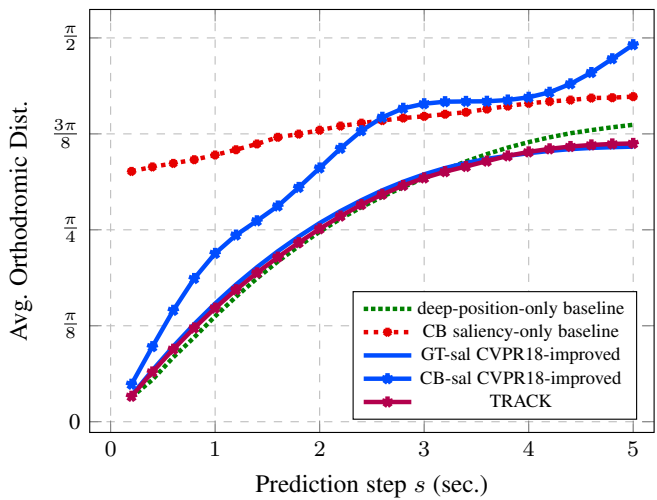}
\caption{Comparison, on the MMSys18 dataset, of TRACK with baselines and both CB-sal CVPR18-improved and GT-sal CVPR18-improved.}
\label{fig:TRACK-avg}
\end{figure}

On the MMSys18 dataset introduced in Sec. \ref{sec:with_GTsal} (with higher user motion, and balanced video categories) and for prediction horizons up to 5 sec., Fig. \ref{fig:TRACK-avg} compares the results of TRACK with both CB-sal CVPR18-improved and GT-sal CVPR18-improved. Indeed, GT-sal CVPR18-improved is considered as a lower-bound on the error of CVPR18, which does not use PanoSalNet (and whose implementation is not available online nor was communicated on request). 
We observe that TRACK outperforms CB-sal CVPR18-improved (as expected), and equates to GT-sal CVPR18-improved, which is remarkable. This confirms the importance of the additional architectural elements of TRACK, able to exploit the (approximated) CB-saliency.

\subsubsection{Comparison with all methods on their original metrics and $H\leq 2.5$ sec.}

For the sake of space, the results of TRACK against all the considered existing methods, on their original metrics and prediction horizons, are presented in Sec. \ref{sec:results_literature}. It can be seen that on every dataset, TRACK (always with CB-saliency) establishes state-of-the-art performance: Fig. \ref{fig:block_NOSSDAV_CVPR}-Left shows that it outperforms MM18 (which also uses PanoSalNet), Table \ref{table:Res_PAMI18} shows that it significantly outperforms PAMI18, as does Table \ref{table:Res_NOSSDAV17} for NOSSDAV17. \new{ChinaCom18 is trained with the leave-one-out strategy, and the dataset is the same as NOSSDAV17. The results of TRACK listed against NOSSDAV17 in Table \ref{table:Res_NOSSDAV17} are thus a lower-bound to TRACK's performance if it were trained with the leave-one-out strategy, already outperforming ChinaCom18 by more than 30\%.} \new{As expected from the answer to Q2 in Sec. \ref{sec:answerA2}, for such short prediction horizons ($H\leq2.5$ sec.), TRACK does not outperform the \textit{deep-position-only baseline}.
Its slightly inferior performance is due to the fact that we did not do any hyperparameter tuning for TRACK, while we did for the \textit{deep-position-only baseline} which is smaller (tuning the number of layers and neurons). When training for $H=5$ sec., the next results in Fig. \ref{fig:eval_avg_alldatasets} and \cite[Sec. VIII]{us_supplemental} show that, for $s\leq 3$ sec., TRACK is similar to or even outperforms the \textit{deep-position-only baseline} for 4 datasets in 5.}

\subsubsection{Exhaustive cross-comparison with all methods on all datasets with the orthodromic distance and $H=5$ sec.}

\begin{figure*}[h]
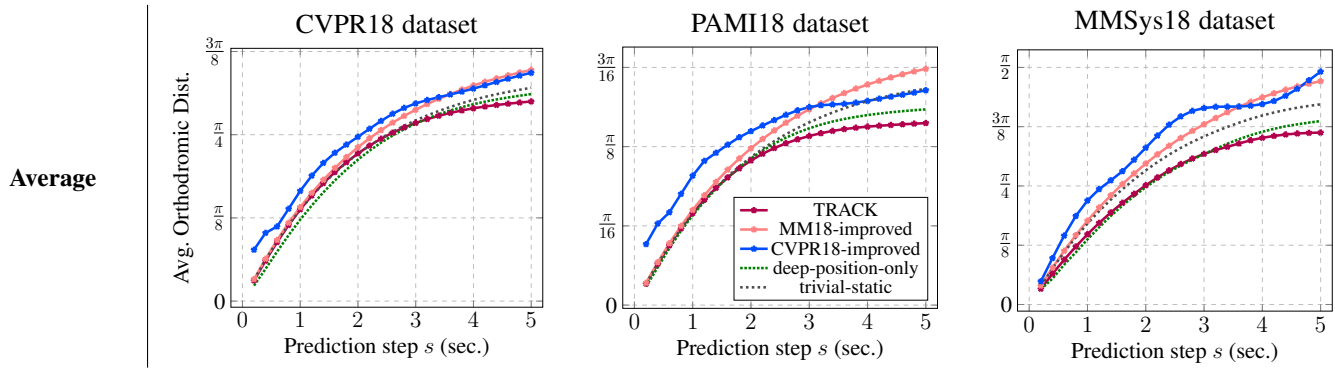

  \centering
\begin{center}
\begin{tabularx}{\textwidth}{c|ccc}
\parbox{0.1\textwidth}{\textbf{\ \ Average}} & \raisebox{-.5\height}{\input{Figures/EvalAverage_CVPR18}} & \raisebox{-.5\height}{\input{Figures/EvalAverage_PAMI18}} & \raisebox{-.5\height}{\input{Figures/EvalAverage_MMSys18}}\\
\end{tabularx}
\end{center}
  \caption{\new{Evaluation results of the methods TRACK, MM18-improved, CVPR18-improved and baselines, averaged over all test videos for the datasets of CVPR18, PAMI18 and MMSys18. Results for the NOSSDAV17 and MM18 datasets are shown in \cite[Fig. 4]{us_supplemental}. Legend and axis labels are the same in all figures.}}
  \label{fig:eval_avg_alldatasets}
\end{figure*}

\new{\textbf{Average results}: Fig. \ref{fig:eval_avg_alldatasets} presents the performance, on all 5 datasets (CVPR18, PAMI18, MMSys18, MM18, NOSSDAV17) of every (re-)implemented method, all with CB-sal: TRACK, CVPR18-improved, MM18-improved, \textit{deep-position-only baseline}, \textit{trivial-static baseline}.
The results are averaged over the videos in the respective test sets made of 42 videos for CVPR18, 16 for PAMI18, 4 for MMSys18, 2 for MM18 and 2 for NOSSDAV17.\\
$\bullet$ For prediction steps $s\geq 3$ sec., TRACK outperforms all methods on all five datasets, except for the NOSSDAV17 dataset where it equates to the best (likely because the \textit{saliency-only baseline} does not outperform the \textit{deep-position-only baseline} on the NOSSDAV17 dataset, as shown in Fig. \ref{fig:A2_NOSSDAV-MM-dataset}).\\
$\bullet$ For $s\leq 3$ sec., TRACK equates to the best method which is the \textit{deep-position-only baseline}, except on the CVPR18 dataset where it has a slightly inferior performance but equates to the other methods.}

\begin{figure*}[h]
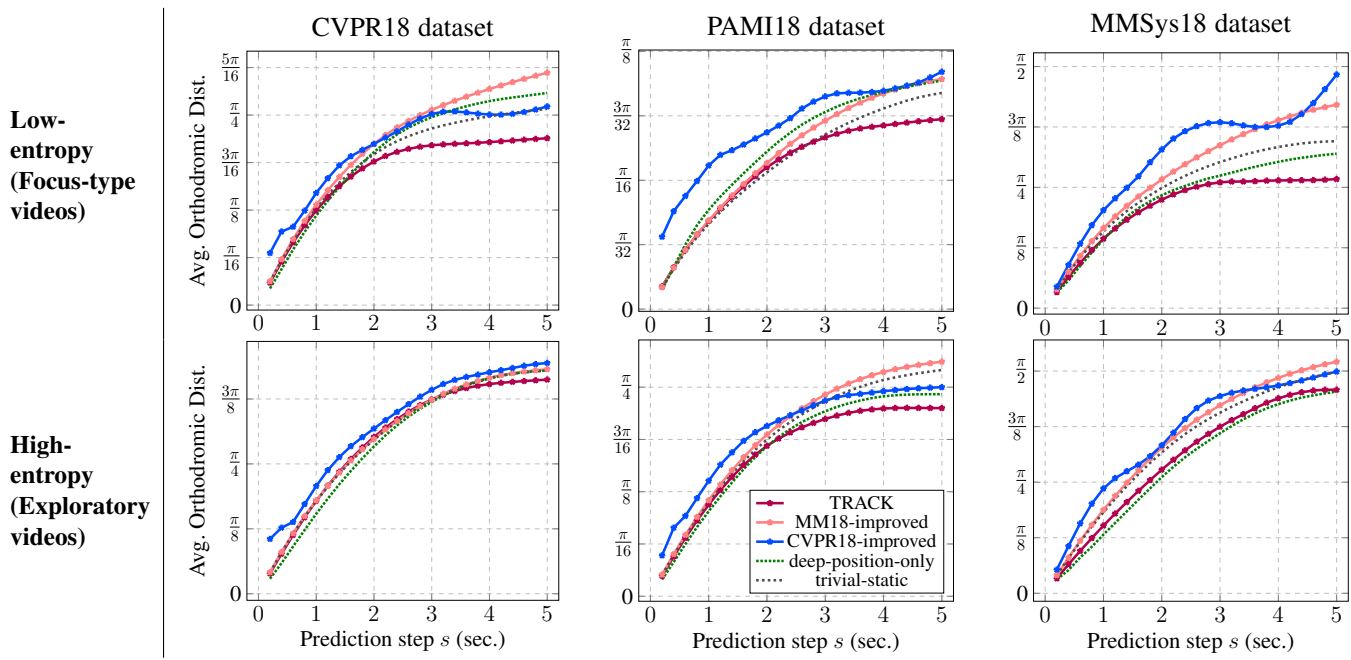

\begin{center}
\begin{tabularx}{\textwidth}{c|ccc}
\parbox{0.1\textwidth}{\textbf{Low-entropy \\ (Focus-type videos)}} & \raisebox{-.5\height}{\input{Figures/LowEntropyCVPR18}} & \raisebox{-.5\height}{\input{Figures/LowEntropyPAMI18}} & \raisebox{-.5\height}{\input{Figures/LowEntropyMMSys18}}\\
\parbox{0.1\textwidth}{\textbf{High-entropy \\ (Exploratory videos)}} & \raisebox{-.5\height}{\input{Figures/HighEntropyCVPR18}} & \raisebox{-.5\height}{\input{Figures/HighEntropyPAMI18}} & \raisebox{-.5\height}{\input{Figures/HighEntropyMMSys18}}\\
\end{tabularx}
\end{center}
\caption{\new{Top row (resp. bottom row): results averaged over the 10\% test videos having lowest entropy (resp. highest entropy) of the GT saliency map. For the MMSys dataset, the sorting has been made using the Exploration/Focus categories presented in Sec. \ref{sec:A2}. Legend and axis labels are the same in all figures.}}
\label{fig:eval_avg_per_entropy}
\end{figure*}

\noindent\new{\textbf{Gains on video categories}: 
The results in Sec. \ref{sec:A2} have shown that the gains that can be expected from a multimodal architecture over the \textit{deep-position-only baseline} are different depending on the video category: whether it is a \textit{focus-type} or an \textit{exploratory} video. The results in Fig. \ref{fig:eval_avg_alldatasets}, averaged over all the videos of a test set, are therefore not entirely representative of the gains. To analyze the gains of TRACK over different video categories, we proceed as follows. First, we only focus on the CVPR18, PAMI18 and MMSys18 datasets to have a sufficient number of videos in the test set. Then, for MMsys18, we group the test videos into a Focus category (with Waterpark and Warship) and an Exploration category (with Portoriverside and Energybiolab), as done in Sec. \ref{sec:A2}. 
Finally for CVPR18 and PAMI18, in order to apply this binary categorization Focus vs Exploratory, we rely on the users behavior. Indeed, the more the users tend to have a focusing behavior, the lower the entropy of the GT saliency map\footnote{The entropy of the 2D map is computed per frame, then averaged over all the frames for $t\geq 6$ sec. to skip the exploratory phase.}. Thus we consider the entropy of the GT saliency map of each video to assign the video to one category or the other. We sort the videos of the test set in increasing entropy, and we represent in Fig. \ref{fig:eval_avg_per_entropy} the results averaged over the bottom 10\% (focus-type videos) and top 10\% (exploratory videos).\\
$\bullet$ On the low-entropy/focus-type videos and for $s\geq 3$ sec., TRACK significantly outperforms the second-best method: by 16\% for PAMI18 to 20\% for both CVPR18 and MMSys18 at $s=H=5$ sec.. TRACK performs similarly or better for $s<3$ sec..\\
$\bullet$ On the high-entropy/exploratory videos, the gains of TRACK are much less significant: TRACK often performs similarly or slightly worse than the \textit{deep-position-only baseline}, yet never degrading significantly away from this baseline, as the other methods do.}
Such results are expected from the observations drawn in Sec. \ref{sec:A2} (Fig. \ref{fig:A2_MMSys-dataset},\ref{fig:A2_PAMI18-dataset},\ref{fig:A2_CVPR18-dataset}) showing that the \textit{saliency-only baseline} does not outperform the \textit{deep-position-only baseline} on exploratory videos.

\begin{figure}[h]
\centering
\hspace*{0.6cm} \textbf{\large CVPR18 dataset}  \hspace*{1.4cm} \textbf{\large MMSys18 dataset}\hfill
\input{Figures/LowEntropy_CVPR_Video072} \hfill \input{Figures/LowEntropy_MMsys_Warship}\hfill
\includegraphics[width=0.49\columnwidth]{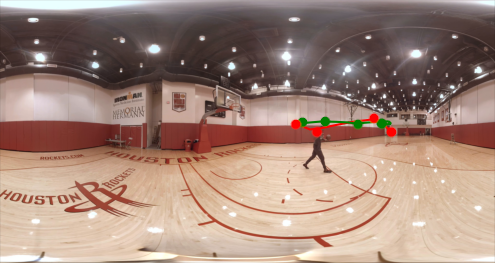} \hfill
\includegraphics[width=0.49\columnwidth]{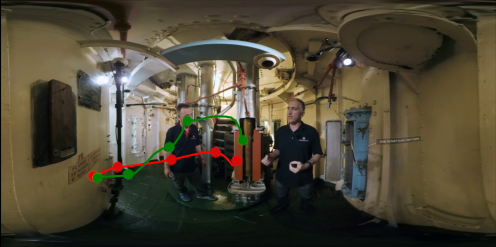}\hfill
\caption{\new{Example of performance on two individual test videos of type Focus. On the frame, the green line represents the ground truth trajectory, and the corresponding prediction by TRACK is shown in red. More examples on other datasets are provided in the supplemental material \cite[Fig. 5]{us_supplemental}.}}
\label{fig:eval_ind_video}
\end{figure}

\noindent\new{\textbf{Qualitative examples}:
In Fig. \ref{fig:eval_ind_video}, we exemplify the results on two low-entropy videos, also showing a representative frame with a user's future trajectory and the prediction of TRACK. More examples are in \cite[Sec. VIII]{us_supplemental}. On focus-type videos, TRACK outperforms significantly the second-best method: by up to 25\% in the examples.
}

\subsection{Ablation study of TRACK}

\begin{figure}[!t]
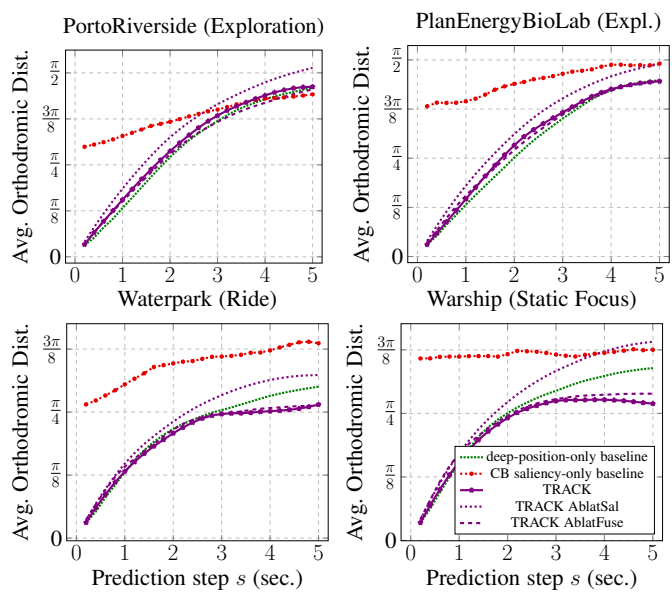

\centering
\input{Figures/Saliency_ContentBasedAblated_PortoRiverside}\hfill
\input{Figures/Saliency_ContentBasedAblated_PlanEnergyBioLab}\hfill
\input{Figures/Saliency_ContentBasedAblated_Waterpark}\hfill
\input{Figures/Saliency_ContentBasedAblated_Warship}\hfill
\caption{Per-video results of TRACK and ablation study. The legend is identical for all sub-figures.}
\label{fig:ablation-videos}
\end{figure}

To confirm the analysis that led us to introduce this new architecture TRACK for dynamic head motion prediction, we perform an ablation study of the additional elements we brought compared to CVPR18-improved: we either replace the RNN processing the CB-saliency with two FC layers (line named AblatSal in Fig. \ref{fig:ablation-videos}), or replace the fusion RNN with two FC layers (line named AblatFuse).

Fig. \ref{fig:TRACK-avg} and \ref{fig:ablation-videos} confirm the analysis in Sec. \ref{sec:analysis5}: the removal of the first extra RNN (not present in CVPR18) processing the saliency input has more impact: AblatSal degrades away from the \textit{deep-position-only baseline} in the first time-steps. The degradation is not as acute as in CVPR18-improved as the fusion RNN can still modulate over time the importance of CB-saliency. However, it seems this fusion RNN cancels most of its input (position and saliency features), as the performance of AblatSal is consistently similar to that of the \textit{trivial-static baseline} (not plotted for clarity).
The AblatFuse line shows that the impact of removing the fusion RNN is less important.\\

\noindent\textbf{Answer to Q4}:
If O1 is solved approximately by providing a saliency estimate obtained from the video content only, the good choice (CVPR18-improved) for Q3 is not sufficient anymore. A RNN dedicated to processing the saliency must be added to prevent the prediction in the first time-steps from degrading away from the \textit{deep-position-only baseline}.
Our new deep architecture, named TRACK, achieves state-of-the-art performance on all considered datasets and prediction horizons.

\section{Discussion}\label{sec:discussion}
It is interesting to note that only a few architectures have been designed in the same way as TRACK, and none for head motion prediction.
Indeed, following up on \cite{jain2016structural}, Sadeghian et al. in \cite{sadeghian2017tracking} proposed a similar architecture to predict a pedestrian's trajectory based on the image of the environment, the past ego trajectory and the trajectories of others. 
Let us also mention that the CVPR18 block is similar to an early architecture proposed for visual question answering in 2015 \cite{refVQA}, and PAMI18 is similar to Komanda proposed in 2016 for autonomous driving \cite{refKomanda}.

This article brings a critical analysis to existing deep architectures aimed to predict the user's head motion in 360\degree\ videos from past positions and video content. As we exhibit the weaknesses of the evaluation scenarios considered by previous works (dataset and competitor baselines), it is important to mention that other such critical analyses have been made for other application domains of deep learning very recently. Indeed, besides Martinez et al. mentioned earlier who showed in \cite{martinez2017human} the weakness of existing architectures for 3D-skeleton pose prediction, Ferrari Dacrema et. al. performed an analysis of recommendation systems in \cite{dacrema2019we}. Not only did they show the difficulty to reproduce the evaluated algorithms, but also that the state-of-the-art methods could not outperform simple baselines.
Similarly, the meta-analysis of Yang et. al. \cite{yang2019critically} for information retrieval, and Musgrave et. al. \cite{musgrave2020metric} for loss functions, show that, contrary to the claims of the authors of multiple recent papers, there has been no actual improvement in several years of proposed neural networks to solve the problem in each of these fields.

In \cite{blalock2020state}, Blalock et. al. show that the difficulty to reproduce, measure and compare the performances of different algorithms makes it difficult to determine how much progress has been made in a field, and this difficulty grows when each work uses different datasets, different performance metrics and different baselines. 
In the present article, we have faced the same difficulties. From an entire reproducible framework \cite{us_gitlab} we have made to enable replication and comparison, we could perform a critical and constructive analysis. 

Our approach and findings are therefore aligned with other critical re-examinations of existing works in other application domains of deep learning. 

\section{Conclusion}\label{sec:conclu}
This article has brought two main contributions. First, we carried out a critical and principled re-examination of the existing deep learning-based methods to predict head motion in 360\degree\ videos, with the knowledge of the past user's position and the video content. We have shown that all the considered existing methods are outperformed, on their datasets and with their test metrics, by baselines exploiting only the positional modality. To understand why, we have analyzed the datasets to identify how and when should the prediction benefit from the knowledge of the content. We have analyzed the neural architectures and shown there is only one whose performance does not degrade compared with the baselines, provided that ground-truth saliency information is provided, and none of the existing architectures can be trained to compete with the baselines over the 0-5 sec. horizon when the saliency features are extracted from the content.
Second, decomposing the structure of the problem and supporting our analysis with the concept of Structural-RNN, we have designed a new deep neural architecture, named TRACK. 
\new{TRACK establishes state-of-the-art performance on all the prediction horizons $H\in[$0 sec.,5 sec.$]$ and all the datasets of the existing competitors. In the 2-5 sec. horizon, TRACK outperforms the second-best method by up to 20\% on focus-type videos, i.e., videos with low-entropy saliency maps.}

The experimental setup and datasets (whose formats we homogenized) of each assessed method and all our codes, are illustrated and provided online at \cite{us_gitlab}. This reproducible framework has already obtained an ACM reproducibility badge \cite{us_ODS}, and allows the community to easily test any predictor.

In future works, we will investigate deep attention mechanisms to refine the time- and space-varying fusion of modalities, as well as consider variational approaches (with VRNN) to also obtain confidence on the prediction, which is crucial for decision-making.

\section*{Acknowledgments}
This work has been partly supported by the French government, through the UCA JEDI and EUR DS4H Investments in the Future projects ANR-15-IDEX-0001 and ANR-17-EURE-0004. This work was partly supported by EU Horizon 2020 project AI4Media, under contract no. 951911 (\url{https://ai4media.eu/}).

\ifCLASSOPTIONcaptionsoff
  \newpage
\fi

\bibliographystyle{IEEEtran}
\bibliography{egbib}

\begin{IEEEbiography}[{\includegraphics[width=1in,height=1.25in,clip,keepaspectratio]{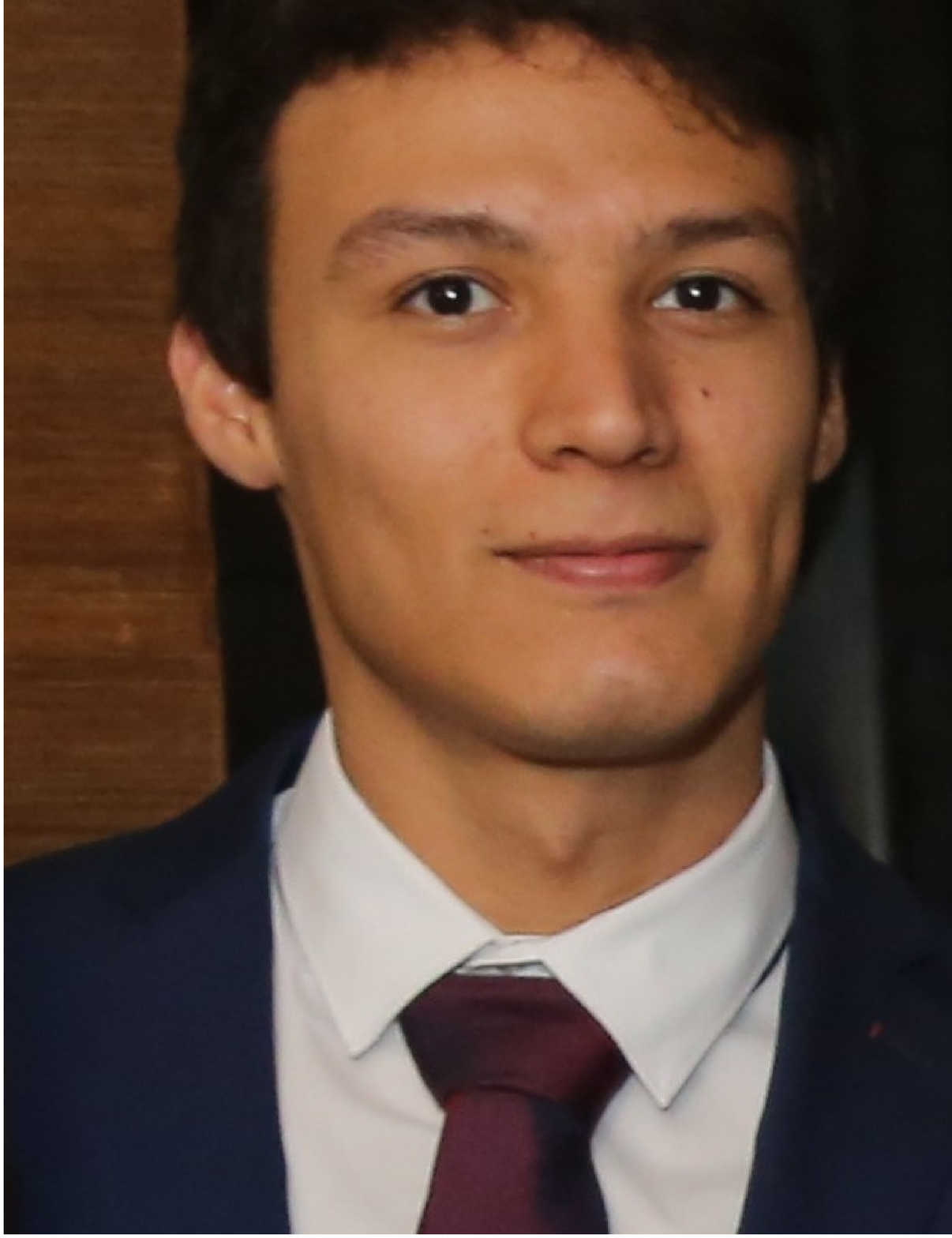}}]{Miguel~Fabi\'an~Romero Rond\'on}
is a PhD student from the Universit\'e C\^ote d'Azur in France since 2017. His PhD thesis titled 'Streaming Virtual Reality: Learning for Attentional Models and Network Optimization,' is in the field of Machine Learning and Networking to identify how to structure the VR content in the cloud and optimally decide what to transmit to maximize the streaming quality. His research interests include Deep Learning, Multimedia Streaming, Virtual Reality and related applications.
\end{IEEEbiography}

\begin{IEEEbiography}[{\includegraphics[width=1in,height=1.25in,clip,keepaspectratio]{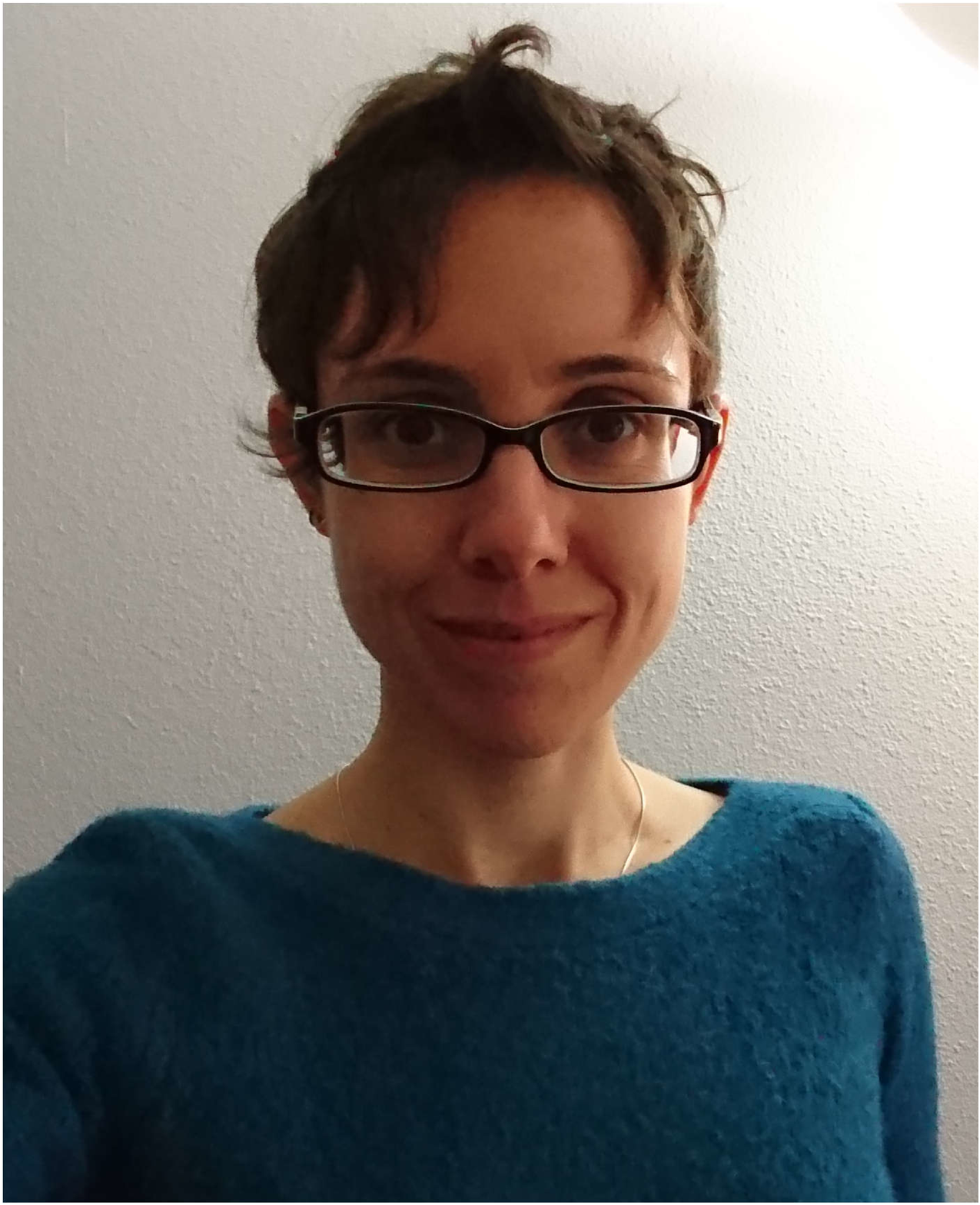}}]{Lucile Sassatelli}
has been an associate professor with Universit\'e C\^ote d'Azur since 2009, and she is a junior fellow of Institut Universitaire de France (IUF). She obtained a PhD from Universit\'e of Cergy Pontoise, France, and her professorial habilitation in 2019. She focuses on the problems of multimedia transmission of virtual and augmented reality (VR and AR), specifically on machine learning-based streaming approaches.
\end{IEEEbiography}

\begin{IEEEbiography}[{\includegraphics[width=1in,height=1.25in,clip,keepaspectratio]{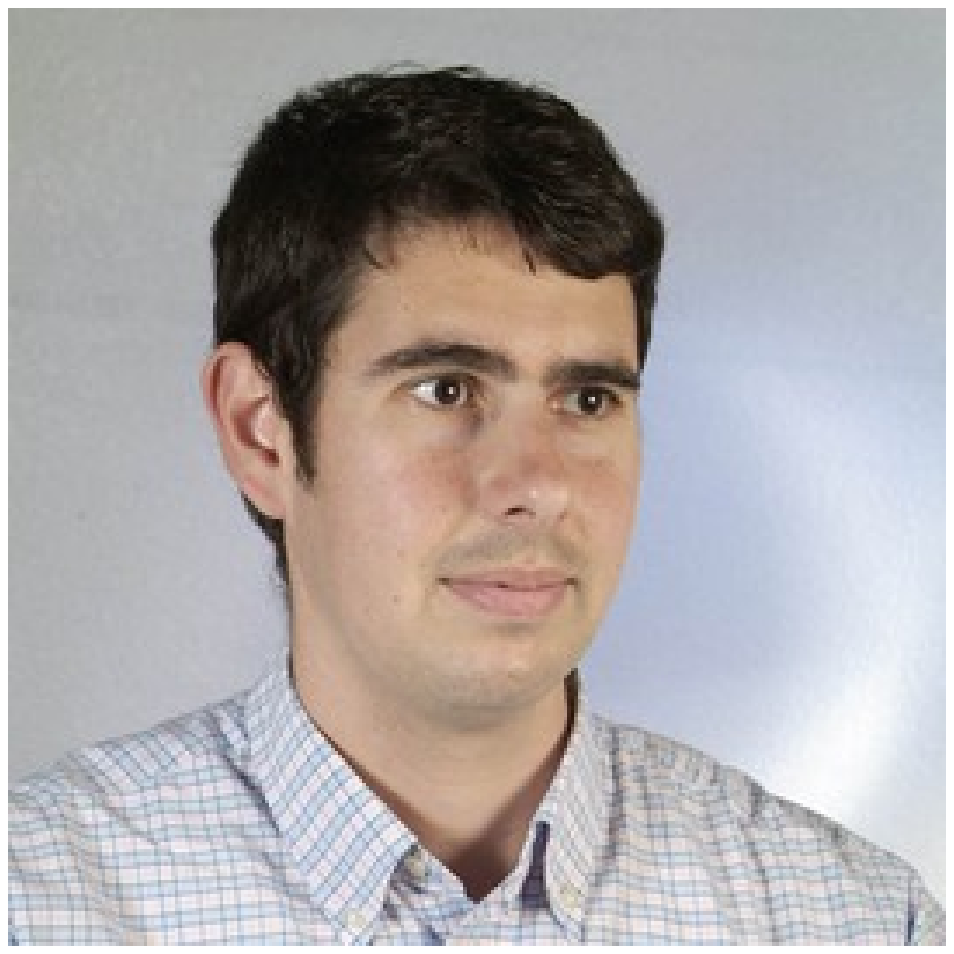}}]{Ram\'on~Aparicio Pardo}
is an associate professor at Universit\'e C\^ote d'Azur (UCA), France, since 2015. He received a MSc and a PhD from Universidad Polit\'ecnica de Cartagena (UPCT), Spain, in 2006 and 2011, respectively. His Ph.D. thesis was distinguished with Telefonica Award for Best Thesis in Networking. His research interests are in optimal design and management of communication networks, and more recently on machine learning-driven network control.
\end{IEEEbiography}

\begin{IEEEbiography}[{\includegraphics[width=1in,height=1.25in,clip,keepaspectratio]{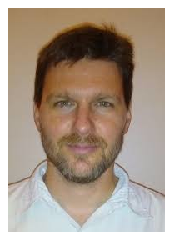}}]{Fr\'ed\'eric Precioso}
received his PhD from Universit\'e Nice Sophia Antipolis in 2004. From 2005 to 2011, he had been an Associate Professor with engineering school ENSEA, Cergy, France. Since 2011, he has been a Full Professor at Universit\'e C\^ote d'Azur (UCA). Since 2018, Prof. Precioso is the Scientific and Program Officer on AI for the National Research Agency. His research interests are on Machine Learning and Deep Learning, specifically for computer vision, audio and linguistics applications, with a strong focus on interpretability and fairness.
\end{IEEEbiography}

\end{document}